\documentclass{article}
\usepackage{ijcai22}

\usepackage{times}
\usepackage{soul}
\usepackage{url}
\usepackage[hidelinks]{hyperref}
\usepackage[utf8]{inputenc}
\usepackage[small]{caption}
\usepackage{graphicx}
\usepackage{amsmath}
\usepackage{amsthm}
\usepackage{bbm}
\usepackage{amsfonts}
\usepackage{amssymb}
\usepackage{booktabs}
\usepackage{subcaption}
\usepackage{algorithm}
\usepackage{algorithmic}
\usepackage{wrapfig}

\usepackage{xcolor}

\title{Online Planning in POMDPs with Self-Improving Simulators}

\author{
Jinke He$^1$
\and
Miguel Suau$^1$\and
Hendrik Baier$^{2}$\and
Michael Kaisers$^{2}$ \And
Frans A. Oliehoek$^1$
\affiliations
$^1$Delft University of Technology, The Netherlands\\
$^2$Centrum Wiskunde \& Informatica, The Netherlands
\emails
\{J.He-4, M.SuaudeCastro, F.A.Oliehoek\}@tudelft.nl,
\{hendrik.baier, michael.kaisers\}@cwi.nl
}

\begin{document}

\maketitle


\newcommand{\E}{\mathop{\mathbb{E}}}

\renewcommand{\S}{\mathcal{S}} 
\newcommand{\A}{\mathcal{A}} 
\newcommand{\T}{\mathcal{T}} 
\renewcommand{\O}{\mathcal{O}} 
\newcommand{\R}{\mathcal{R}} 
\renewcommand{\H}{\mathcal{H}}

\newcommand{\local}{\mathit{local}}
\newcommand{\M}{\mathcal{M}}
\newcommand{\G}{\mathcal{G}}
\newcommand{\GM}{\mathcal{M}_\mathtt{global}}
\newcommand{\IALM}{\mathcal{M}_\mathtt{IALM}}
\newcommand{\SL}{\mathcal{S}^\mathtt{IALM}}
\newcommand{\TL}{\mathcal{T}^\mathtt{IALM}}
\newcommand{\bL}{b_0^\mathtt{IALM}}

\newcommand{\simGM}{\mathcal{G}_\mathtt{global}}
\newcommand{\simIALM}{\mathcal{G}_\mathtt{IALS}}
\newcommand{\simIALMp}{\mathcal{G}_\mathtt{IALS}^\theta}

\newcommand{\Slocal}{S^{\mathtt{local}}}
\newcommand{\slocal}{s^{\mathtt{local}}}
\newcommand{\sglobal}{s^{\mathtt{global}}}
\newcommand{\Snonlocal}{S^{\neg \mathtt{local}}}
\newcommand{\snonlocal}{s^{\neg \mathtt{local}}}
\newcommand{\SIALM}{S^{\mathtt{IALM}}}
\newcommand{\sIALM}{s^{\mathtt{IALM}}}

\newcommand{\D}{\mathcal{D}} 
\newcommand{\piexplore}{\pi_{\mathtt{explore}}}

\newcommand{\KLD}[2]{D_{\mathrm{KL}} \left( \left. \left. #1 \right|\right| #2 \right) }

\newcommand{\aip}{\hat{I}_\theta}

\newcommand{\Ysrc}{Y^{\mathtt{src}}}
\newcommand{\Xdest}{X^{\mathtt{dest}}}
\newcommand{\ysrc}{y^{\mathtt{src}}}
\newcommand{\ssrc}{s^{\mathtt{src}}}
\newcommand{\Ssrc}{S^{\mathtt{src}}}
\newcommand{\xdest}{x^{\mathtt{dest}}}
\newcommand{\Xnondest}{X^{\neg \mathtt{dest}}}
\newcommand{\xnondest}{x^{\neg \mathtt{dest}}}

\newcommand{\simLocal}{\mathcal{G}_{\mathtt{local}}}
\newcommand{\simSI}{\mathcal{G}_{\mathtt{SI}}}

\newcommand{\SIALS}{S^{\mathtt{IALS}}}
\newcommand{\sIALS}{s^{\mathtt{IALS}}}

\newcommand{\tree}{\mathtt{tree}}
\newcommand{\pitree}{\pi^{\tree}}

\newcommand{\env}{\mathtt{env}}

\newcommand{\sumCE}{\mathcal{L}^{CE}}
\newcommand{\sumE}{\mathcal{L}^{E}}
\newcommand{\sumKL}{\mathcal{L}^{KL}}
\newcommand{\nCE}{n^{CE}}
\newcommand{\nE}{n^{E}}
\newcommand{\nKL}{n^{KL}}

\newcommand{\nGlobal}{N({\simGM} )}
\newcommand{\nIALS}{N ({\simIALMp} )}

\newcommand{\aMeta}{a^{\mathtt{meta}}}
\newcommand{\cMeta}{c^{\mathtt{meta}}}

\newcommand\blfootnote[1]{%
  \begingroup
  \renewcommand\thefootnote{}\footnote{#1}%
  \addtocounter{footnote}{-1}%
  \endgroup
}

\begin{abstract}
    How can we plan efficiently in a large and complex environment when the time budget is limited?
    Given the original simulator of the environment, 
    which may be computationally very demanding, 
    we propose to learn online an approximate but much faster simulator that improves over time. 
    To plan reliably and efficiently while the approximate simulator is learning, 
    we develop a method that
    adaptively decides which simulator to use for every simulation, 
    based on a statistic that measures the accuracy of the approximate simulator. 
    This allows us to use the approximate simulator to replace the original simulator for faster simulations  
    when it is accurate enough under the current context, 
    thus trading off simulation speed and accuracy. 
    Experimental results in two large domains show that when integrated with POMCP, 
    our approach allows to plan with improving efficiency over time. 
\end{abstract}

\section{Introduction}

Decision making under uncertainty is one of the key problems in artificial intelligence \cite{Kaelbling1998,spaan2012partially}. 
Online planning methods such as POMCP \cite{Silver2010} and DESPOT \cite{NIPS2013_c2aee861} have become popular as they enable highly efficient decision making by focusing the computation on the current context. 
However, these methods rely heavily on fast simulators that can rapidly perform a large number of Monte Carlo simulations. 
Unfortunately, many real-world domains are complex and thus slow to simulate, which impedes real-time online planning.     

To mitigate the cost of simulation, researchers have explored learning surrogate models that are computationally less expensive \cite{grzeszczuk1998neuroanimator,buesing2018learning,chitnis2020learning}. 
However, one disadvantage of these methods is that they learn a model of the \emph{entire} environment, which may be unnecessary and is often a difficult task.  
A recent approach by He \emph{et al.}~\shortcite{NEURIPS2020_2e6d9c60} builds on the framework of influence-based abstraction to overcome this problem: 
They propose to exploit the structure of the environment via a so-called influence-augmented local simulator (IALS), which uses a lightweight simulator to capture the local dynamics that surround the agent and learns an influence predictor $\aip$ that accounts for the interaction between the local dynamics and the rest of the environment. 

However, there are three main limitations of this "two-phase" paradigm, in which a simulator is learned offline and then used as-is for online simulation and planning.
First, no planning is possible until the offline learning phase finishes, which can take a long time.
Second, the separation of learning and planning raises the question of what data collection policy should be used during training to ensure accurate online prediction during planning.  We empirically demonstrate that when the training data is collected by a uniform random policy, the learned influence predictors can perform poorly during online planning, due to distribution shift. 
Third, completely replacing the original simulator with the approximate one after training implies a risk of poor planning performance in certain situations, which is hard to detect in advance.

In this work, we aim to overcome these drawbacks by investigating whether we can learn the influence predictor $\aip$ used in an IALS \emph{online, without any pretraining}. As the quality of such an IALS with an untrained influence predictor would be poor initially, we investigate if during online planning, we can selectively use the learned IALS and the accurate but slow global simulator (GS) to balance simulation speed and accuracy. To address this major challenge, we propose a simulator selection mechanism to choose between the GS and IALS based on the latter's accuracy under the current context, which we estimate online with the simulations of the GS. This enables us to use the fast IALS when it is sufficiently accurate, while reverting back to the GS otherwise.

Our experiments reveal that as the influence predictor becomes more accurate, the simulator selection mechanism starts to use the IALS more and more often, which allows for high-accuracy simulations with increasing speed. Planning with such \emph{self-improving simulators} speeds up decision making (given a fixed number of simulations) and increases task performance (given a fixed time budget), without pretraining the influence predictor $\aip$. Moreover, we find that influence predictors that are trained with online data, i.e., coming from online simulations of the GS, can significantly outperform those trained with offline data, both in terms of online prediction accuracy and planning performance.\blfootnote{Extended version of this paper with supplementary material is available at \url{https://arxiv.org/abs/2201.11404}.}

\section{Background}

\paragraph{POMDPs}
A Partially Observable Markov Decision Process (POMDP) \cite{Kaelbling1998} is a tuple $\M = (\S, \A, \Omega, \T, \R, \O, \gamma)$ where $\S$, $\A$ and $\Omega$ represent the state, action and observation space. 
At every time step $t$, the  decision making agent takes an action $a_t$, 
which causes a transition of the environment state from $s_t$ to $s_{t+1} \sim \T(\cdot | s_t, a_t )$, and receives 
a reward $r_t = \R(s_t, a_t)$ and a new observation $o_{t+1} \sim \O(\cdot | s_{t+1}, a_t)$. 
A policy $\pi$ of the agent maps an action-observation history $h_t = (a_0, o_1, \ldots, a_{t-1}, o_t)$ to a distribution over actions.
The value function $V^{\pi}(h_t)$ denotes the expected discounted return $\E_{\pi}[\sum_{k=0}^{\infty} \gamma^k r_{t+k}|h_t]$ when a history $h_t$ is observed and a policy $\pi$ is followed afterwards. 
The value for taking action $a_t$ at $h_t$ and following $\pi$ thereafter is captured by the action-value function $Q^{\pi}(h_t, a_t)$.
The optimal action-value function denotes the highest value achievable by any policy $Q^{*}(h_t, a_t) = \max_{\pi} Q^{\pi}(h_t, a_t)$.
\paragraph{POMCP}
Classical POMDP techniques \cite{Kaelbling1998,spaan2012partially} scale moderately as they explicitly represent the exact beliefs over states. To improve scalability, 
POMCP \cite{Silver2010} performs online planning based on Monte Carlo Tree Search ~\cite{Browne2012} to focus the computation on the current context $h_t$, 
while applying particle filtering to avoid representing the full exact belief.

To make a decision after observing history $h_t$, POMCP repeatedly performs Monte Carlo simulations to incrementally build a lookahead search tree with a generative simulator $\mathcal{G}$. Each node in the tree represents a simulated history with visit count and average return maintained for every action. POMCP starts simulation $i$ by sampling a particle $s^{\mathcal{G}}_t$ (a possible current environment state) from the root node, which is then used to simulate a trajectory $\tau_i$. The simulated actions are chosen by the UCB1 algorithm \cite{Auer2002} to balance exploration and exploitation when the simulated histories are represented in the tree, and otherwise by a random rollout policy. 
After simulation, the statistics of the tree nodes are updated with the simulated trajectory and exactly one node is added to the tree to represent the first newly encountered history.
The whole planning process can be seen as approximating the local optimal action-value function $\hat{Q}^{*}(h_t,\cdot)$.
As the result of planning, the action that maximizes the average return $a_t {=} \arg \max_{a} \hat{Q}^{*}(h_t, a)$ is returned. 

\paragraph{Influence-Augmented Local Simulators}
\begin{figure}
  \begin{minipage}[b]{0.49\linewidth}
  \centering
    \includegraphics[page=1,width=0.78\linewidth]{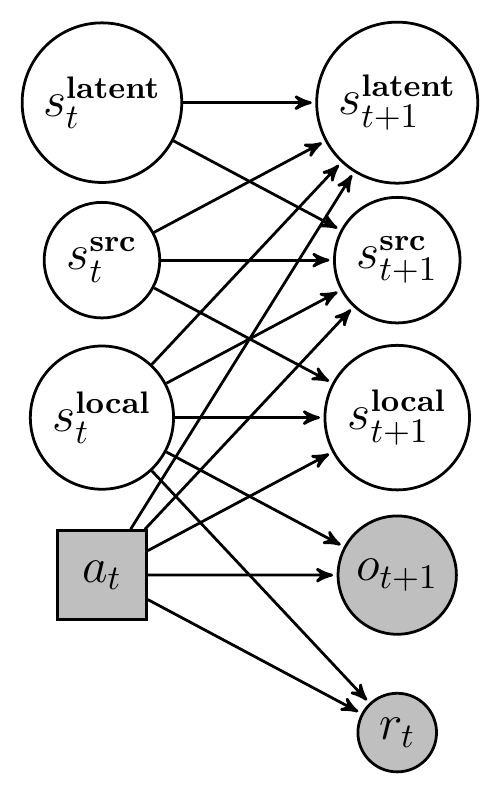}
  \end{minipage}
  \begin{minipage}[b]{0.49\linewidth}
    \centering
    \includegraphics[page=2,width=0.78\linewidth]{PGM.pdf}
  \end{minipage}
  \caption{Two-stage Dynamic Bayesian Network (2DBN); observable variables (grayed), and control input (square). Full model (left), and the model of IALS (right) with local history dependent influence source state distribution (colored).}
  \label{fig:illustration}
\end{figure}

While POMCP has led to great improvement in planning efficiency, it does not address the problem of representing a very large POMDP. For concise representations, Factored POMDPs \cite{hansen2000dynamic} formulate a state $s_t$ as an assignment of state variables and exploit conditional independence between them to compactly represent the transition and reward models with a so-called Two-stage Dynamic Bayesian Network (2DBN) \cite{Boutilier1999}, as illustrated in Figure~\ref{fig:illustration}. Such compact representations enable us to represent very large domains with thousands of state variables, but sampling all of them at every simulation is still too costly.  By using influence-based abstraction (IBA) \cite{Oliehoek21JAIR} to abstract away state variables that do not directly affect the agent's observation and reward, it is possible to simulate and plan more efficiently in large POMDPs \cite{NEURIPS2020_2e6d9c60}. 

Consider the left 2DBN in Figure~\ref{fig:illustration}, a model that can be readily used for simulation, which, however, might be too slow if there are many state variables. Fortunately, in many problems, the agent only directly interacts with a small part of the environment. Formally, we define the \emph{local state} $\slocal_t \subseteq s_t$ as the subset of state variables that do directly affect the observation and reward of the agent. Then, the \emph{influence source state} $\ssrc_t{\subseteq} s_t {\setminus} \slocal_t$ is defined as the state variables that directly affect the local state. By combining the observation and reward models with the \emph{local transition function} that models the transition of the local state $\slocal_{t+1} {\sim} \T^{\mathtt{local}}(\cdot | \slocal_t, \ssrc_t, a_t)$, an abstract \emph{local model} can be constructed, which concisely models the agent's observation and reward with the only external dependency being the influence source state $\ssrc_t$. Figure~\ref{fig:illustration} (Right) illustrates the influence-augmented local simulator (IALS), which consists of a simulator of the local model $\simLocal$ that can be extracted from the 2DBN, and an influence predictor $I$. The influence predictor models the influence source distribution $I(\ssrc_t | d_t)$ conditioned on the local states and actions $d_t {=} ( \slocal_0, a_0, \slocal_1, \ldots, a_{t-1}, \slocal_t )$.  Note that the latent state variables $s_t {\setminus} (\slocal_t \bigcup \ssrc_t)$ are abstracted away in this local simulator, compared to the global simulator that simulates all the state variables in the 2DBN, thereby boosting simulation efficiency (see Table~\ref{table:simulators} for comparison).

\begin{table*}[t]
    \small
    \centering
    \begin{tabular}{|l|l|l|l|}
    \hline
    Simulator                                                       & Feature  &   State & Simulation \\ \hline
    the global simulator $\simGM$ & accurate \& slow & $s_t$ & $s_{t+1}, o_{t+1}, r_t \sim \simGM(s_t, a_t)$       \\ \hline
    the IALS $\simIALMp = (\simLocal, \aip)$ & approximate \& fast & $s^{\mathtt{IALS}}_t = (\slocal_t, d_t)$ &
      \begin{tabular}{@{}c@{}} $\ssrc_t \sim \aip(\cdot | d_t) $ \\ $\slocal_{t+1}, o_{t+1}, r_t \sim \simLocal(\slocal_{t}, \ssrc_t, a_t) $ \end{tabular} \\ \hline
    \end{tabular}
    \caption{Comparison between the global simulator (GS) and the influence-augmented local simulator (IALS).}
    \label{table:simulators}
    \vspace{-1em}
\end{table*}

Although the IALS is proven to be unbiased with an exact influence predictor $I$, i.e., optimal planning performance can still be achieved when replacing the global simulator with the IALS~\cite{Oliehoek21JAIR}, the required inference to obtain $I$ is generally intractable. To address this, He \emph{et al.}~\shortcite{NEURIPS2020_2e6d9c60} propose a two-phase approach that first learns an approximate influence predictor with recurrent neural networks (RNNs) ~\cite{Cho2014} offline and then uses it as-is as part of the IALS, to replace the global simulator for faster simulations. Experimental results show that this leads to an increase in simulation efficiency, which improves the online planning performance under limited planning budget. Algorithm~2 (appendix) summarizes the procedure of applying this approach. 

\section{Planning with Self-improving Simulators}


To address the limitations of the two-phase approach mentioned before (the need for pre-training, the distribution mismatch between states visited during training and planning, and the risk of catastrophic planning performance implied by replacing the original simulator completely with the approximate one after training), we introduce a new approach that 
1) starts planning from the start while collecting the data to train the surrogate model, 
 2) trains the influence predictor with this data, which is now more pertinent to planning, and 
 3) reverts back to the global simulator $\simGM$ when the IALS is considered inaccurate for the current context (i.e., the current history and future trajectories we are likely to reach).
 Additionally, by switching back to $\simGM$ when the IALS is inaccurate, we provide more training to improve it precisely for those inaccurate contexts. 
 As such the IALS will improve over time, and be used more frequently, such that overall simulation efficiency improves over time. 
 Therefore, we refer to our approach as a \emph{self-improving simulator (SIS)}.

We propose and test the integration of these ideas in POMCP, although the principle is transferable to 
other planning methods such as DESPOT that also iteratively perform simulations.
Algorithm \ref{alg:planSIS} outlines our approach. 
The planning starts with an arbitrarily initialized influence predictor $\aip$,
which is combined with the 
local simulator $\simLocal$ that captures the local dynamics $(\T^{\local}, \O, \R)$
to form 
an IALS $\simIALMp = (\simLocal, \aip)$. 
%
For every simulation, a simulator is selected between $\simGM$ and $\simIALMp$ to produce a new trajectory $\tau_i$ (line~\ref{line:planSISSelect}).  This trajectory is processed as in regular POMCP to update statistics and expand one new tree node (line~\ref{line:planSISupdate}), regardless which simulator was used. Note that this means that the constructed tree carries information from both simulators.
In line~\ref{line:planSISstore}, however, only when $\tau_i$ comes from $\simGM$, we extract and store training data from $\tau_i$. Action selection (line~\ref{line:planSISact}) is the same as in regular POMCP.


The data stored, denoted as $\D{=}\{ (d_k, \ssrc_k) \}$ for any $0{\leq}k{\leq}\H{-}1$, can then be used as a replay buffer to further improve $\aip$ regularly. For example, Algorithm~\ref{alg:planSIS} suggests doing so at the end of each real episode. 
To train the approximate influence predictor, parameterized as a recurrent neural network, we follow \cite{NEURIPS2020_2e6d9c60} and treat it as a sequential classification task where the cross entropy loss is minimized by stochastic gradient descent \cite{Ruder2016}:
\begin{align*}
  \mathcal{L}(\theta ; \D) = - \E_{d_k, \ssrc_k \sim \D} \log{\aip(\ssrc_k | d_k)}.
\end{align*}

Overall, we expect to mainly select the global simulator for earlier simulations, since $\aip$ is not yet very accurate. Over time, we expect the IALS to become more accurate, and thus more frequently used. However, global simulations are needed to assess the accuracy of the IALS, thus leading to a complex exploration/exploitation problem.

\begin{algorithm}[tb]
\small
\caption{Planning with the Self-improving Simulator}
\label{alg:planSIS}
\begin{algorithmic}[1] 
\STATE initialize the influence predictor $\aip$
\FOR{every real episode} 
\FOR{every time step $t$} 
\FOR{every simulation $i=0,\ldots$} 
\STATE \label{line:planSISSelect} 
select either $\simGM$ or $\simIALMp$ and simulate trajectory $\tau_i$
\STATE \label{line:planSISupdate} 
update the search tree with $\tau_i$
\STATE 
\label{line:planSISstore}
if $\simGM$ generated $\tau_i$, add training data to $\D$
\ENDFOR
\STATE 
\label{line:planSISact}
take greedy action, and prune tree with new observation
\ENDFOR
\STATE 
\label{line:planSIStrain}
train the influence predictor $\aip$ on $\D$ for $N$ steps
\ENDFOR
\end{algorithmic}
\end{algorithm}

\section{Online Simulator Selection}
\label{section:selection}

The major challenge in our approach is how to select between the simulators online for fast \emph{and} accurate simulations. Since the decision quality of IALS pivots on the influence source distribution aligning with what is knowable about the actual random influences in the global simulator, our solution begins with an information-theoretic accuracy measure of $\aip$, 
which we then show can be estimated online. 
Subsequently, we describe how to address the exploration/exploitation problem when selecting simulators online with the UCB1 algorithm. 

\subsection{IALS Accuracy Measure}
To address the question of measuring the accuracy of the IALS, one idea is to measure the Kullback–Leibler (KL) divergence between the true influence source distribution and the one predicted by the influence predictor. This is motivated by theory~\cite{Congeduti21AAMAS}, which shows that the maximum KL divergence $D_{KL}(I(\Ssrc_k | d_k) || \aip(\Ssrc_k | d_k))$ of any history of local states and actions $d_k$ can be used to bound the value loss. 
However, our accuracy measure should not only consider the accuracy of $\aip$ at the current simulated step but should also take the upcoming simulated steps into account. 

Formally, at every real time step $t$ given the action-observation history $h_t$, we denote the (partial) search tree at the beginning of POMCP simulation $i$ as $\tree_i$ and the tree policy as $\pi^{\tree_i}$. For every simulated time step $k{\geq} t$, we define $P_k(D_k | h_t, \pi^{\tree_i})$ as the distribution over local histories $D_k$ at that time step \footnote{We use the uppercase and lowercase letters to distinguish between random variables and their realizations.}. That is, for a local history $d_k {=} (\slocal_0, a_0, \slocal_1, \ldots, a_{k-1}, \slocal_{k})$, $P_k(d_k | h_t, \pi^{\tree_i})$ defines the probability of the local history at time step $k$ being $d_k$, if the policy $\pi^{\tree_i}$ is executed from time step $t$, knowing that the history at time step $t$ is $h_t$. Then, the expected KL divergence between the true influence source distribution $I$ and the approximate one given by $\aip$ is defined as:
\begin{align*}
  \mathcal{L}_i^k \triangleq \E_{d_k \sim P_k(\cdot | h_t, \pi^{\tree_i})} \KLD{I(\Ssrc_t | d_k) }{ \aip(\Ssrc_t | d_k)}
\end{align*}
Averaging the expected KL divergence over all simulated time steps, we have 
$  \mathcal{L}_i \triangleq \frac{1}{\H-t} \sum_{k=t}^{\H-1} \mathcal{L}_i^k. $
%
The intuition here is that if we are now at real time step $t$ with history $h_t$, and have performed $i{-}1$ POMCP simulations, then $\mathcal{L}_i$ measures the expected average KL divergence between the true influence source distribution and the one approximated by $\aip$, with the local history $d_k$ under the distribution induced by drawing a new POMCP simulation from an IALS with an exactly accurate influence predictor $I$.

\subsection{Estimating the Accuracy}
While we do not have access to the true influence source distribution, i.e., $I(\cdot | d_k)$, we can use the simulations from the global simulator to construct an estimate of the accuracy.

For every simulated step $t{\leq}k{\leq}\H\text{-}1$ during simulation $i$,
 \begin{align*}
  \mathcal{L}_i^k 
  &= \E_{d_k \sim P_k} [\underbrace{H(I(\Ssrc_t | d_k), \aip(\Ssrc_t | d_k))}_{\textstyle \text{cross entropy}\mathstrut}{-}\underbrace{H(I(\Ssrc_t | d_k)}_{\textstyle \text{entropy}\mathstrut})].
\end{align*}
For the cross entropy part, it is possible to show the expectation of it equals 
$\E_{d_k, s_k \sim P_k(\cdot | h_t, \pi^{\tree_i})} - \log{\aip(s^{src}_k | d_k)} $, with full derivation in Appendix~A.1.
This means that this term can be estimated from samples from the global simulator, as long as we keep track of the local history $d_k$ during simulation.

Unlike cross entropy, we cannot directly estimate entropy from samples due to the lack of an unbiased estimator for entropy \cite{paninski2003estimation}. Here we choose to use the entropy of the same distribution but instead conditioned on the global state, as a lower bound of the entropy to estimate (see full derivation in Appendix~A.2):
 \[ 
 \E_{d_k}
 \left[ H(I(\Ssrc_k | d_k)) \right]  
 \geq 
 \E_{s_{k-1}, a_{k-1}}  
 \left[ H(S^{src}_k | s_{k-1}, a_{k-1}) \right],
  \] 
  where both expectations are with respect to  $P_k(\cdot | h_t, \pitree)$.
This inequality holds intuitively because the global state $s_k$, as a Markovian signal, contains more information than the local history $d_k$, on the random variable $\Ssrc_k$. The entropy $H(S^{src}_k | s_{k-1}, a_{k-1})$ can be directly computed with the 2DBN, which we assume is given. Importantly, the use of a lower bound on the entropy means that we estimate an upper bound on the KL divergence, making sure we do not theoretically underestimate the inaccuracy of the IALS simulations.

Overall this constructs an upper bound on the expected average KL divergence between the true influence source distribution and the one approximated by $\aip$, for simulation $i$:
\begin{align*}
  \mathcal{L}_i 
  & \leq \frac{1}{\H-t} \sum_{k=t}^{\H-1} \E_{d_k, s_k \sim P(\cdot | h_t, \pitree)} -\log{\aip(s^{src}_k | d_k)} \\ 
  &\; \; \; \; \; \;\; \; \; \; \; \; \; \;\; \;- \E_{s_{k-1}, a_{k-1} \sim P(\cdot | h_t, \pitree) } H(S^{src}_k | s_{k-1}, a_{k-1}) 
\end{align*}

As mentioned, the quantity above can be estimated with samples from the global simulator by augmenting the state $s_t$ with the history of local states and actions $d_t$. Specifically, to perform a POMCP simulation at time step $t$, we start by sampling a possible pair $(s_t, d_t)$ from the root node. For every simulated step $k$, we first perform the simulation with the global simulator $s_{k+1}, r_k, o_{k+1} \sim \simGM(s_k, a_k)$ and then let $d_{k+1} = d_k a_k \slocal_{k+1}$. We denote the simulated trajectory as $\tau_i = (s_t, d_t, a_t, s_{t+1}, r_t, o_{t+1}, \ldots, a_{\H-1}, s_{\H}, r_{\H-1}, o_{\H})$. The empirical average KL divergence is then:
\begin{align*}
  l_i = \frac{1}{\H-t} \sum_{k=t}^{\H-1} [- \log{\aip(\ssrc_k |d_k) } - H(\Ssrc_k | s_{k-1}, a_{k-1})]
\end{align*}


\subsection{Deciding between the Simulators}

Finally, we want to use the estimated accuracy to decide whether to use the IALS or the global simulator for a simulation. However, alluded to before, the simulations from the GS will not only be used for planning but also for estimating the accuracy of the IALS, and this will need to be done every time we plan for a decision: Even if in the past we determined that $\aip$ is accurate for the subtree of the current history $h_t$, it is possible that due to learning for other histories the accuracy for this $h_t$ has decreased. Moreover, due to the constant updating of the search tree during a planning step, the tree policy also changes, which can also invalidate past accuracy estimates. As such, we face an ongoing exploration (assessing accuracy) and exploitation (of accurate $\aip$) problem.

To address this, we propose to model the question of which simulator to use for the $i$-th simulation as a bandit problem, and apply the UCB1 algorithm \cite{Auer2002}. In the following, we define the values of using the global simulator $\simGM$ and the IALS $\simIALMp$ for every simulation $i$:
\begin{align*}
  &V_i^{\simIALMp} = \hat{\mathcal{L}} + c^{\mathtt{meta}} \sqrt{\frac{\log(N^{\simIALMp})}{i}} \\
  &V_i^{\simGM} = -\lambda +  c^{\mathtt{meta}}\sqrt{\frac{\log(N^{\simGM})}{i}} 
\end{align*}
Here,
$N^{\simGM}$ and $N^{\simIALMp}$
are the number of calls to the GS and the IALS, respectively.
$\hat{\mathcal{L}}$ is an average of the empirical average KL divergence, and 
$\lambda$ 
quantifies the
extra computation cost of using the global simulator for a simulation, compared to the IALS. In practice, it is treated as a hyperparameter as it also reflects how willingly the user may sacrifice simulation accuracy for efficiency. 
Note that even though UCB1 is theoretically well understood, bounding the sample complexity here is impossible as both the history-visit distribution and the accuracy of the IALS can continually change.

\section{Empirical Analysis}
In this section, we first evaluate the main premise of our approach: \emph{can selecting between a global simulator and an online learning IALS lead to a self-improving simulator?} Such improvement can manifest itself in faster decision making (when fixing the number of simulations per real time step), or in better performance (when fixing the time budget for each real time step). We investigate both. 
We also 
compare our on-line learning approach to the existing two-phase approach.

\paragraph{Experimental Setup}
We perform the evaluation on two large POMDPs introduced by \cite{NEURIPS2020_2e6d9c60}, the Grab A Chair (GAC) domain and the Grid Traffic Control (GTC) domain, of which descriptions
can be found in Appendix~C.1. 
In all planning experiments with self-improving simulators, we start with an IALS that makes use of a completely untrained $\aip$, implemented by a GRU; after every real episode it is trained for $64$ gradient steps with the accumulated data from the global simulations.
The results are averaged over $2500$ and $1000$ individual runs for the GAC and GTC domains, respectively. Further details 
are provided in Appendix~C.

\begin{figure*}[t]
    \centering
    \begin{subfigure}[b]{0.245\textwidth}
        \centering
        \includegraphics[width=\textwidth]{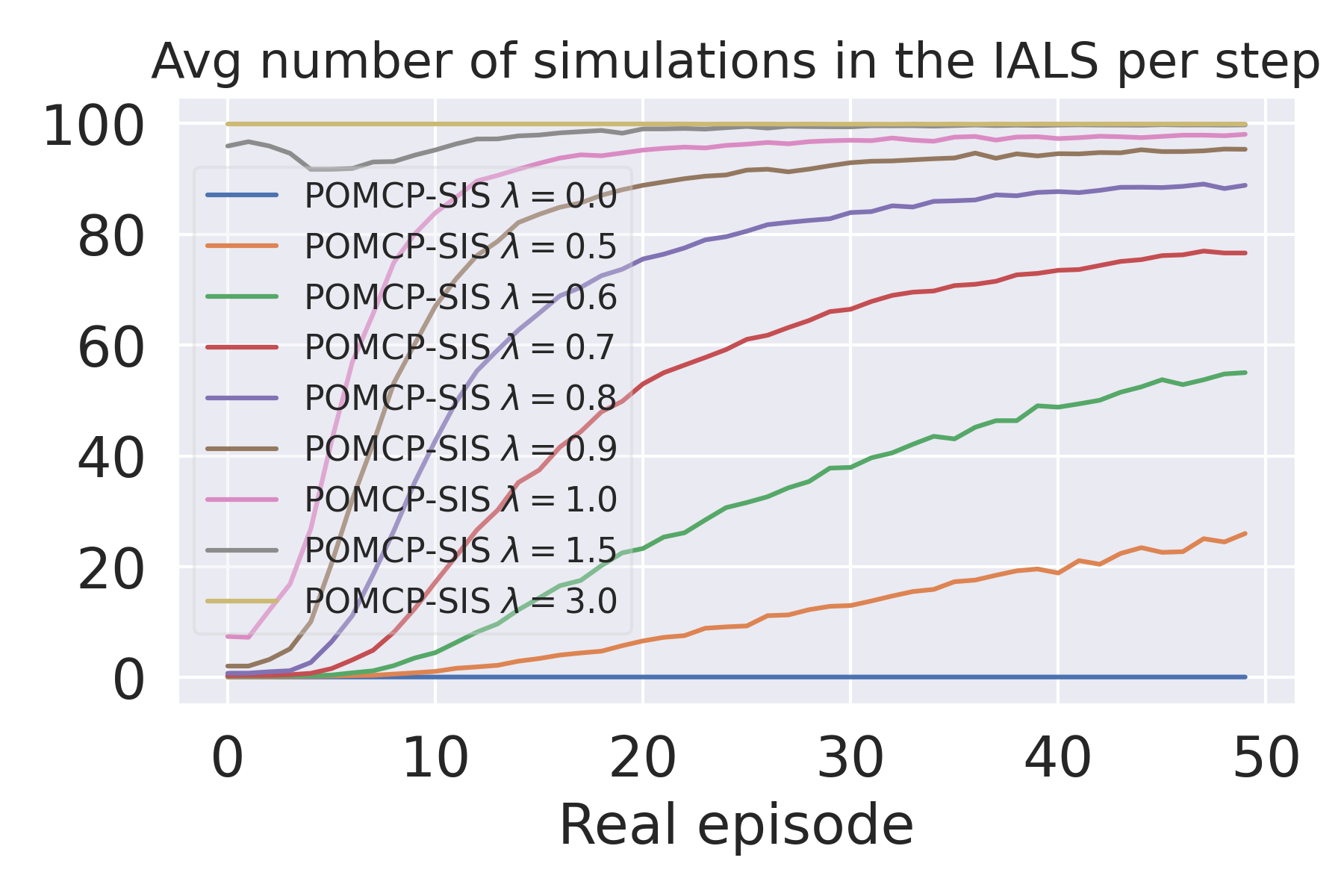}
        \caption{}
    \vspace{-2mm}
        \label{fig:fixed-sim-sim}
    \end{subfigure}
    \begin{subfigure}[b]{0.245\textwidth}
        \centering
        \includegraphics[width=\textwidth]{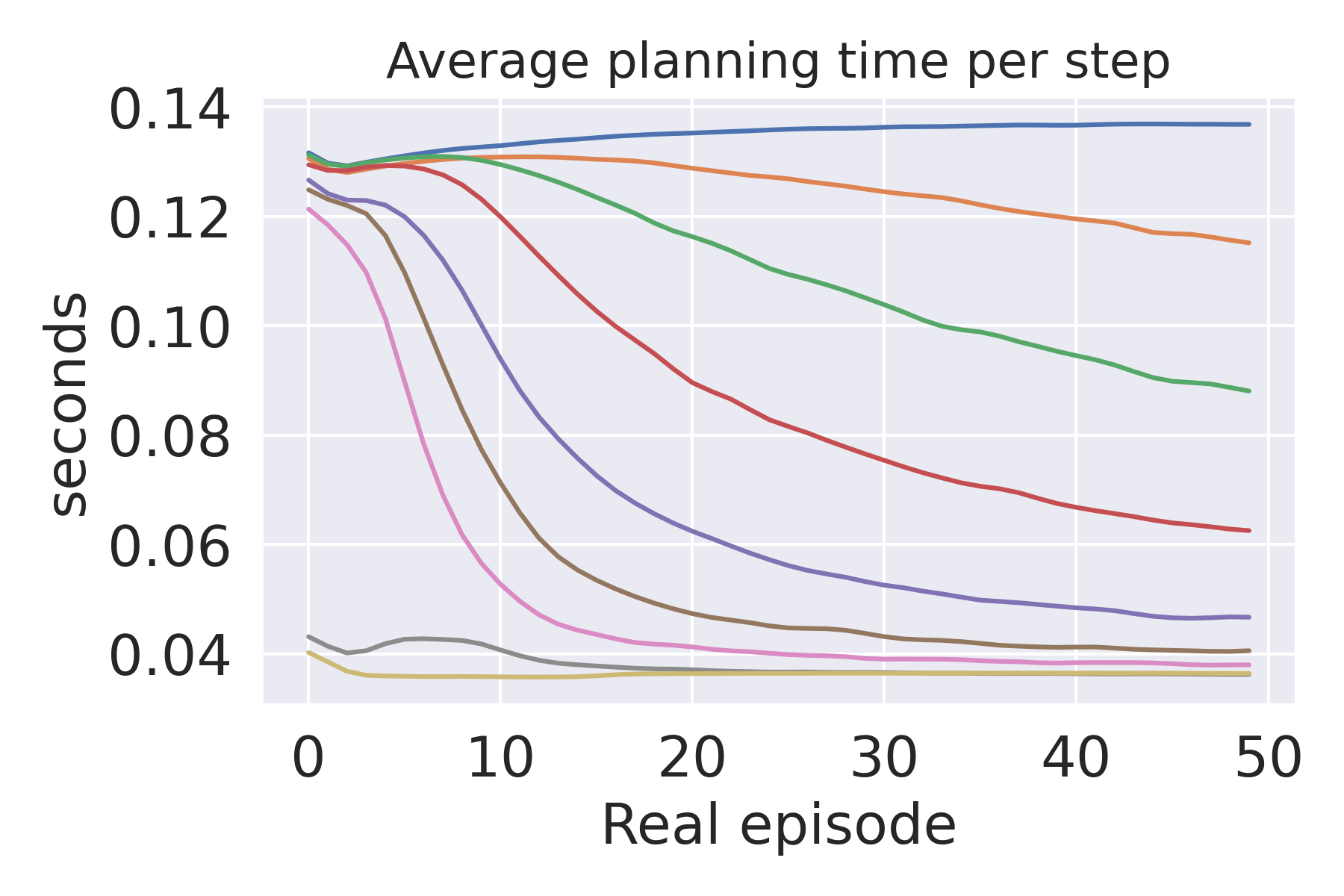}
        \caption{}
    \vspace{-2mm}
        \label{fig:fixed-sim-time}
    \end{subfigure}
    \begin{subfigure}[b]{0.245\textwidth}
        \centering
        \includegraphics[width=\textwidth]{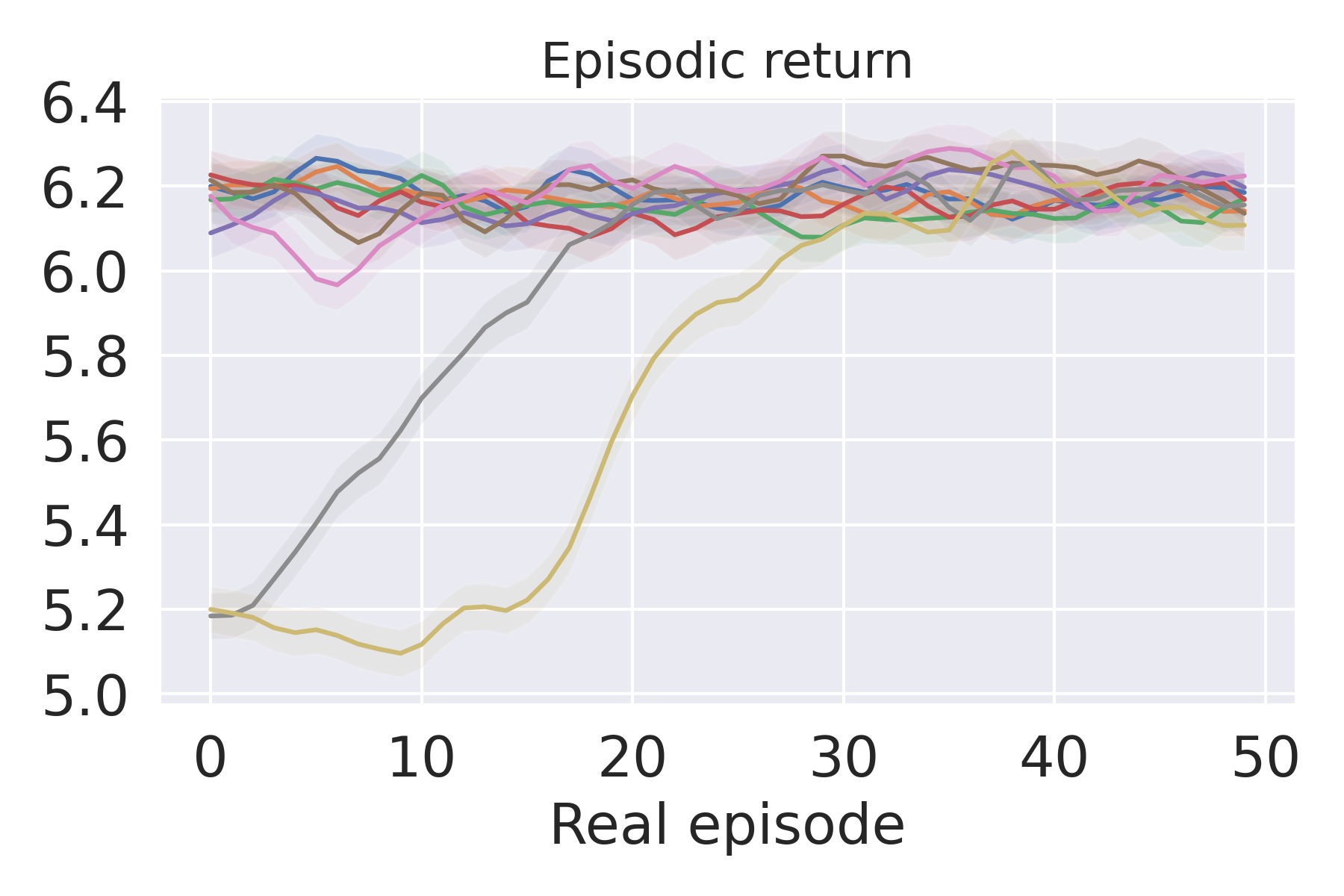}
        \caption{}
    \vspace{-2mm}
        \label{fig:fixed-sim-return}
    \end{subfigure}
    \begin{subfigure}[b]{0.245\textwidth}
        \centering
        \includegraphics[width=\textwidth]{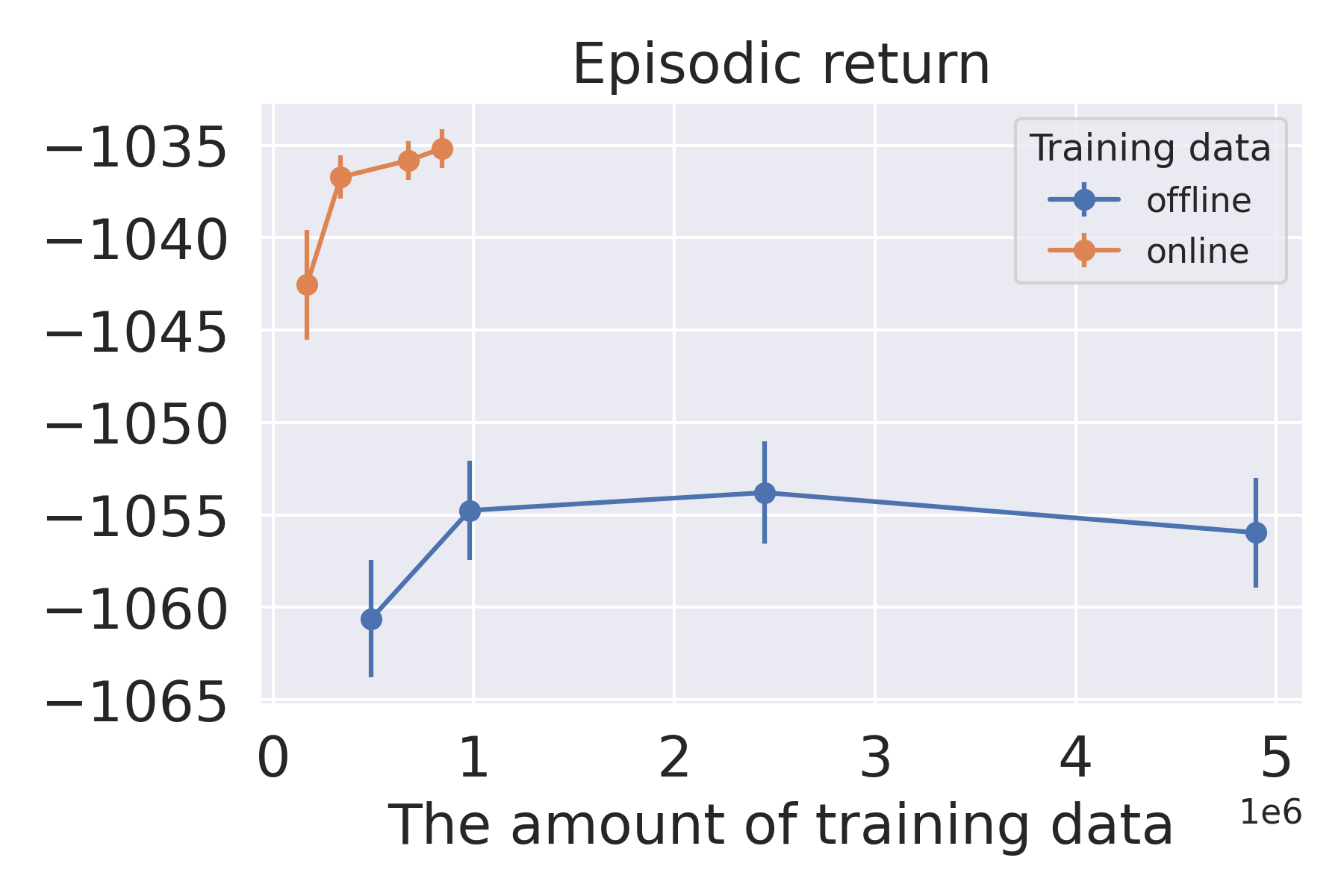}
        \caption{}
    \vspace{-2mm}
        \label{fig:offline-comparison-return}
    \end{subfigure}
    \caption{(a-c) Simulation controlled planning results for grab a chair. Planning time reduces without significant drop in return for small $\lambda$'s. (d) Time controlled planning results for grid traffic control, with IALSs that make use of influence predictors trained on offline (from a uniform random policy) and online (from self-improving simulator with $\lambda{=}{0.7}$) data. This experiment is repeated for $20$ times.}
    \label{fig:simulation-controlled-results}
    \vspace{-2mm}
\end{figure*}

\begin{figure*}[t]
    \centering
    \begin{subfigure}[b]{0.49\textwidth}
        \centering
        \includegraphics[width=.49\textwidth]{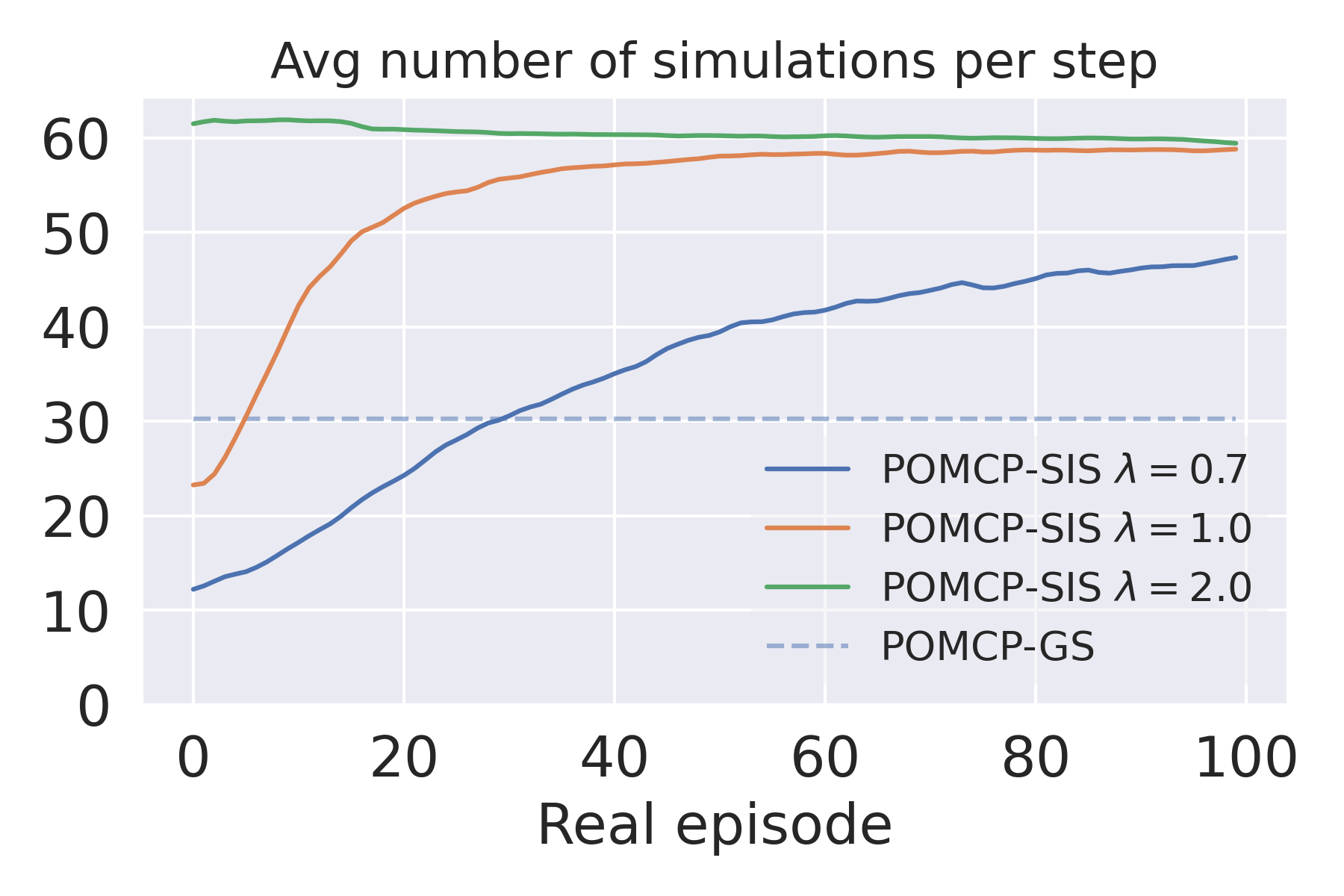}
        \includegraphics[width=.49\textwidth]{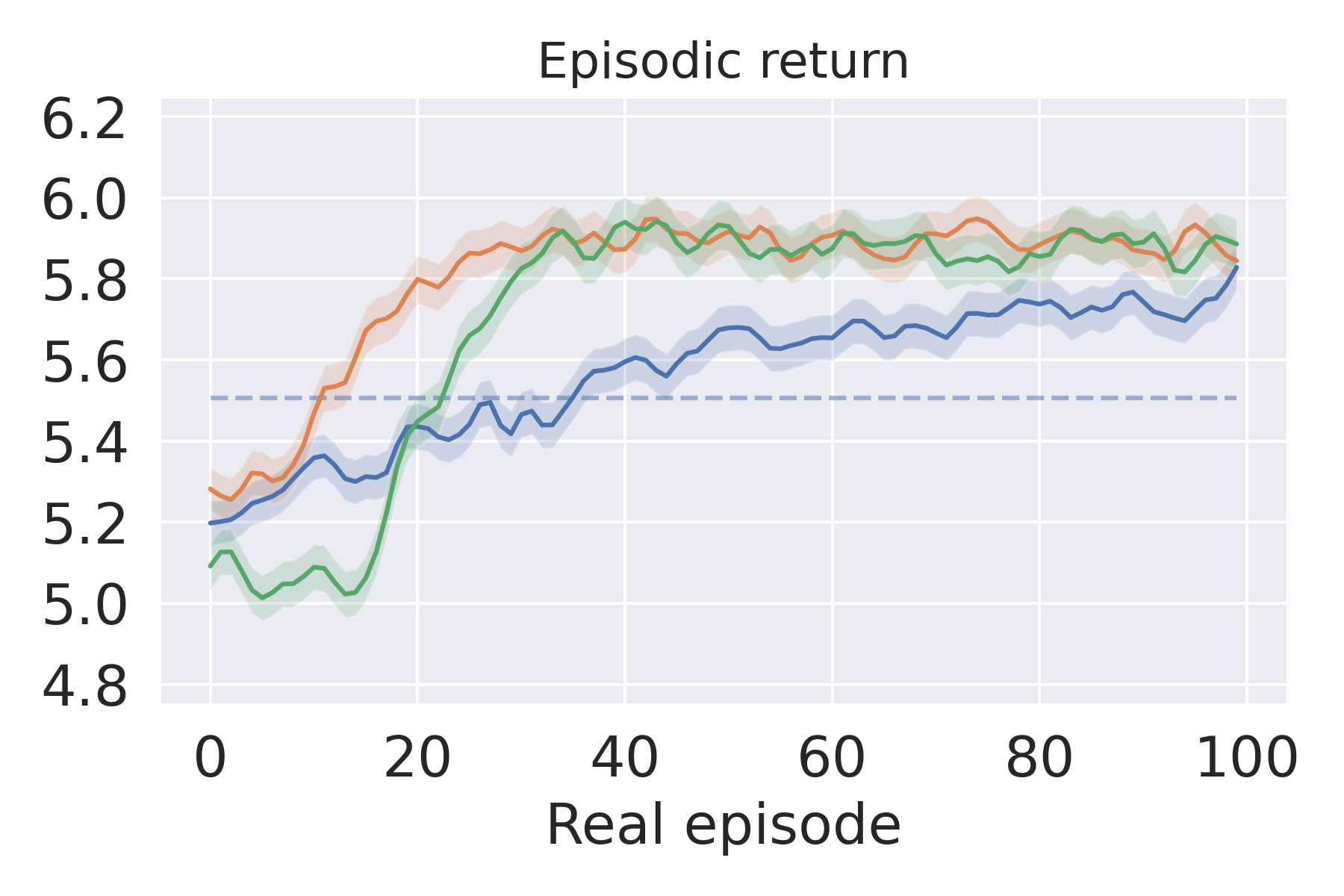}
        \caption{Grab a chair results.}
        \label{fig:real-time-GAC}
    \vspace{-2mm}
    \end{subfigure}
    \begin{subfigure}[b]{0.49\textwidth}
        \centering
        \includegraphics[width=.49\textwidth]{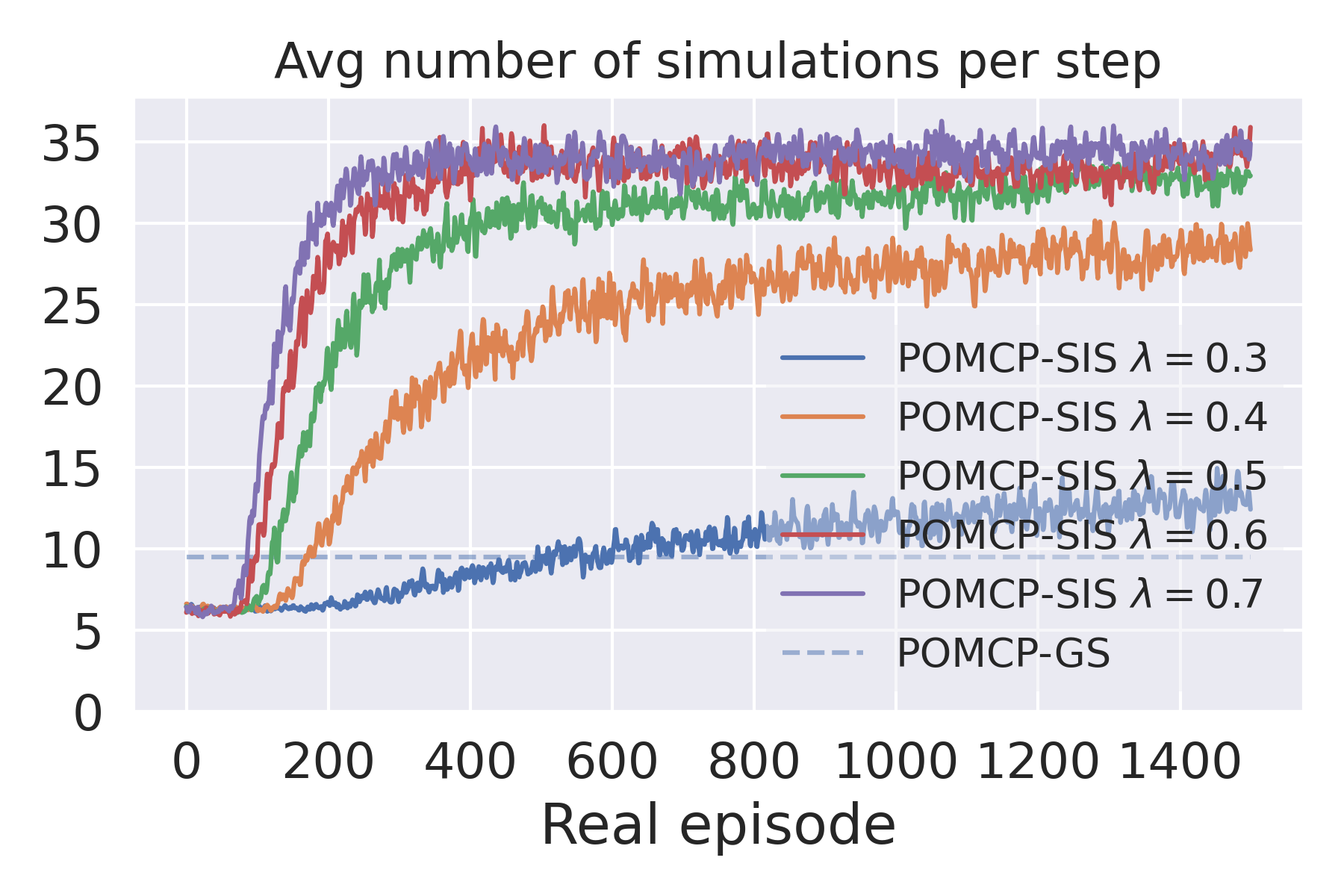}
        \includegraphics[width=.49\textwidth]{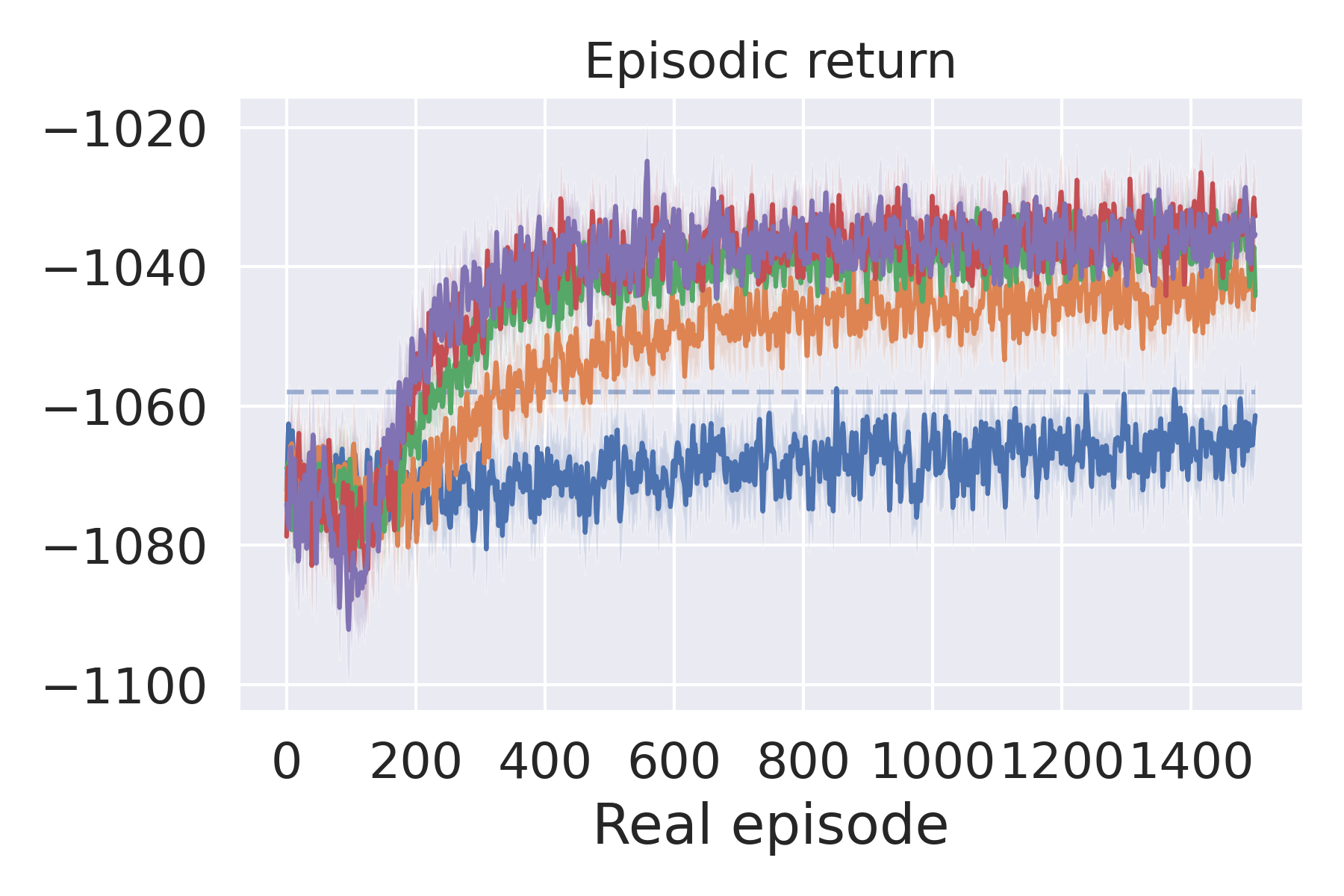}
        \caption{Grid traffic control results.}
        \label{fig:real-time-GTC}
    \vspace{-2mm}
    \end{subfigure}
    \caption{Time controlled planning results. Performance improves because of increasingly accurate IALS and more simulations. }
    \label{fig:real-time-results}
    \vspace{-2mm}
\end{figure*}

\paragraph{Online Planning with Fixed Number of Simulations}
In this experiment, we 
fix the number of POMCP simulations to $100$ per planning step. We ask two questions with this experiment: 1) can planning with self-improving simulators enable increasing planning efficiency? and 2) how does the choice of the hyperparameter $\lambda$ affect the performance of planning? 
Since traffic domains have fixed time constraints, we only evaluate on GAC.
In addition, GAC as the simpler domain allows us to explore many different settings for lambda $\lambda$.

Figure \ref{fig:fixed-sim-sim} shows how many of the 100 simulations are performed using the fast IALS (on average over all real time steps in the episode) for different $\lambda$.
Remember that higher $\lambda$ implies the planner is more willing to sacrifice the simulation accuracy for efficiency.
For moderate $\lambda$s, there is a clear trend that the use of IALS is increasing over time, due to the increasing accuracy of the approximate influence predictor $\aip$, 
which is confirmed by the reduced training loss provided in Appendix~D.
This translates into the increasing planning speed shown in Figure \ref{fig:fixed-sim-time}.%
\footnote{Planning time with $\lambda{=}0.0$ grows slowly due to the practical issue of memory fragmentation, which can be fixed with the memory pool technique from David Silver's implementation of POMCP.} 
Note that large $\lambda$s like $1.5$ and $3.0$ produce very fast planning directly from the first episode. However, as shown in Figure \ref{fig:fixed-sim-return}, the price for that is poor performance at the earlier episodes,
because they begin by heavily using the IALS with untrained influence predictors. 
Moreover, Figures 2b and 2c show that planning time for $\lambda=1.0$ becomes comparable 
to  $\lambda=1.5$ and $\lambda=3.0$ after roughly 10 episodes, while having much better performance at that point.
This can be explained by the lack of training data caused by not using the global simulator often enough.
Importantly, it seems that there is no clear sign of decreasing performance for moderate $\lambda$s, which supports our key argument that in this setting, our approach enables planning with increasing speed without 
sacrificing task performance. 

\paragraph{Real-Time Online Planning}

In this experiment, we evaluate the performance of our approach under limited time budget. On the one hand, the IALS is an approximate model that introduces errors, certainly when the accuracy of its $\aip$ is still poor in early episodes. On the other hand, using the faster IALS means that we can do many more simulations per real time step. As such, we investigate the ability of our proposed simulator selection mechanism to \emph{dynamically} balance between these possible risks and benefits of the IALS.

We perform planning experiments in both the grab a chair domain and the grid traffic control domain, allowing $1/64$ and $1/16$ seconds for each decision correspondingly, with results shown in Figure \ref{fig:real-time-results}. 
We also compare against the baseline that only uses the global simulator and plot their performance as dotted lines. 
We see that there is an overhead with using the self-improving simulators, as shown by the differences in the starting number of simulations. This is the cost of the simulator selection method when estimating the KL divergence with the approximate influence predictor $\aip$, which can potentially be reduced by a faster implementation of the GRUs. This gap is also expected to diminish in even larger domains.


In the grab a chair domain,  we see that the self-improving simulators can perform more simulations for later episodes.
This is expected: for later episodes the  accuracy of the approximate influence predictor $\aip$ improves, 
which means that the IALS is selected more frequently, thus enabling more simulations per real time step on average.
The figure also shows that this larger number of planning simulations translates to better performance: for appropriate values of $\lambda$, after planning for a number of real episodes, the performance increases to a level that planning with the global simulator alone cannot reach.
We stress that, in line with the results in the previous subsection, the increase in return is not \emph{only} explained by doing more simulations: the quality of the $\aip$ does matter, as is clearly demonstrated by $\lambda=2.0$ in GAC: even though it performs many simulations per step from episode 1, its performance only surpasses that of the global simulator baseline after 20 episodes.
Overall, these results show 
that self-improving simulators enables planning with increasing efficiency, and can outperform using an exact global simulator without pre-training an influence predictor.

\paragraph{Further comparison to the two-phase approach}
Another limitation of the two-phase approach is: the offline learned influence predictors may generalize poorly when making online predictions due to the distribution shift issue caused by training with data collected by an exploration policy. We investigate if this is a real issue that can affect performance and if learning with online data, as in our approach, can help. 

To evaluate the two-phase approach, we reuse the real-time planning setting in GTC. Following \cite{NEURIPS2020_2e6d9c60}, we start by collecting a number of datasets of varying sizes with a uniform random policy from the global simulator. We refer to these datasets as the offline data. For comparison, we also construct a number of datasets with different sizes from the replay buffer of the self-improving simulator (with $\lambda{=}{0.7}$), after planning for $1500$ episodes. We refer to these datasets as the online data, because the data is generated during actual planning. Then, different influence predictors are trained on these datasets in an offline phase, before they are used as part of the IALSs for online planning, with results shown in Figure~\ref{fig:offline-comparison-return}.
While it seems that more offline training data can be helpful, overall the performance of the influence predictors that are trained with online data completely dominates those that are trained with offline data: with much less online data, we can achieve a level of performance that is impossible with much more offline data. 
Figure~8 in Appendix shows that this indeed might be caused by a distribution shift issue: when evaluated on a test dataset that is collected by running POMCP on the global simulator, the influence predictors being trained on the offline data can have an increasing test loss. Moreover, training on the online data from the self-improving simulator results in a much lower test loss. As such, we can conclude that there is indeed a distribution shift issue with the two-phase approach that can harm the planning performance, which is addressed in our approach.

\section{Related Work}

Surrogate models have been widely used in many applications, such as aerodynamic design prediction \cite{pehlivanoglu2012aerodynamic}, spacecraft path planning \cite{Peng20AST} and satellite reconfiguration \cite {chen2020satellite}. It has been proposed as a general approach to constraint optimization problems \cite{audet2000surrogate}. 

There is a strong connection between our work and multi-fidelity methods \cite{kennedy2000predicting,peherstorfer2018survey}, which combine models with varying costs and fidelities to accelerate outer-loop applications such as optimization and inference. In essence, the simulator selection mechanism in our approach does the same: It selects between a slow high-fidelity global simulator and a fast low-fidelity IALS, to accelerate planning. In our case however, the fidelity of the IALS may change over time due to the training of the influence predictor, and also depends on the current context. While our simulator selection mechanism is based on an accuracy measure that is theoretically inspired \cite{Congeduti21AAMAS}, future work could incorporate ideas from multi-fidelity methods more explicitly. 


There is also a body of work that applies abstraction methods inside MCTS \cite{hostetler2014state,jiang2014improving,anand2015asap,anand2015novel,anand2016oga,bai2016markovian}. 
Such methods perform abstraction on the MCTS search tree: they aggregate search tree nodes, thus reducing the size of search tree. This is complementary to our technique, which speeds up the simulations themselves.

\section{Conclusion}
In this paper, we investigated two questions. First, given a global simulator (GS) of a complex environment formalized as a factored POMDP, 
can we learn online a faster approximate model, i.e., a so-called IALS? Second, can we balance between such a  learning IALS and the GS for fast \emph{and} accurate simulations to improve the planning performance? 

The answers to these questions lead to a new approach called $\emph{self-improving simulators}$, which selects between the online learning IALS and the GS for every simulation, based on an accuracy measure for IALS that can be estimated online with the GS simulations. As the accuracy of the influence predictor $\aip$ would be poor initially, the GS will be used mainly for simulations in earlier episodes, which can then be used to train $\aip$. This leads to increasing simulation efficiency as the IALS becomes more accurate and is used more often, which improves planning performance (time and return) over time. 

Our approach has three main advantages against the two-phase approach, which trains the influence predictor offline before using it for planning \cite{NEURIPS2020_2e6d9c60}. First, planning can start without pre-training. Second, learning $\aip$ online prevents the distribution shift issue that occurs when transferring the offline learned $\aip$ to make online predictions. Third, selecting the simulator online enables reverting back to the GS when the IALS is not sufficiently accurate for the current context. Future work can investigate different ways of estimating the IALS accuracy that make less use of the expensive GS.

\section*{Acknowledgements}
\newlength{\tmplength}
\setlength{\tmplength}{\columnsep}
This project received funding from the European Research 
\begin{wrapfigure}{r}{0.3\columnwidth}
    \vspace{-10pt}
    \hspace{-12pt}
    \includegraphics[width=0.32\columnwidth]{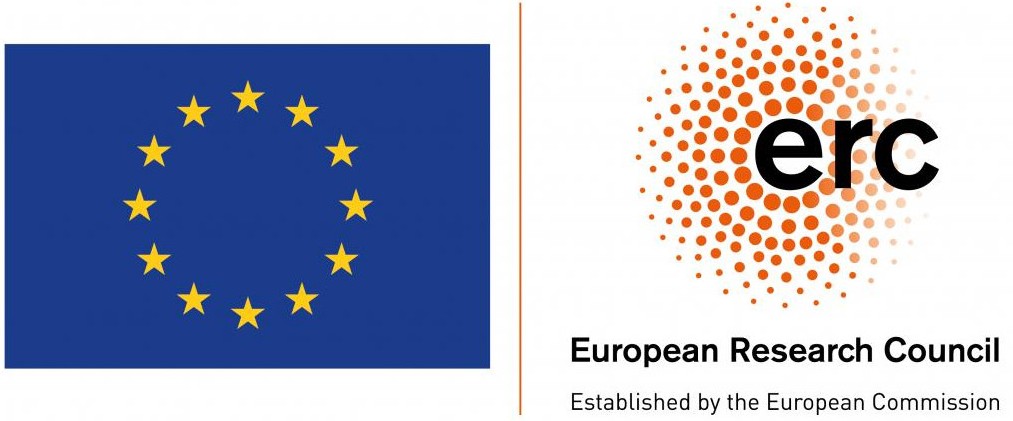}
\end{wrapfigure}
Council (ERC) under the European Union's Horizon 2020 research and innovation programme (grant agreement No.~758824 \textemdash INFLUENCE).
\setlength{\columnsep}{\tmplength}

\bibliographystyle{named}
\bibliography{ijcai22}

\begin{thebibliography}{}

\bibitem[\protect\citeauthoryear{Anand \bgroup \em et al.\egroup
  }{2015a}]{anand2015novel}
Ankit Anand, Aditya Grover, and Parag Singla.
\newblock A novel abstraction framework for online planning.
\newblock In {\em AAMAS}, 2015.

\bibitem[\protect\citeauthoryear{Anand \bgroup \em et al.\egroup
  }{2015b}]{anand2015asap}
Ankit Anand, Aditya Grover, Parag Singla, et~al.
\newblock {ASAP-UCT}: Abstraction of state-action pairs in {UCT}.
\newblock In {\em IJCAI}, 2015.

\bibitem[\protect\citeauthoryear{Anand \bgroup \em et al.\egroup
  }{2016}]{anand2016oga}
Ankit Anand, Ritesh Noothigattu, Parag Singla, et~al.
\newblock {OGA-UCT}: On-the-go abstractions in {UCT}.
\newblock In {\em ICAPS}, 2016.

\bibitem[\protect\citeauthoryear{Audet \bgroup \em et al.\egroup
  }{2000}]{audet2000surrogate}
Charles Audet, J~Denni, Douglas Moore, Andrew Booker, and Paul Frank.
\newblock A surrogate-model-based method for constrained optimization.
\newblock In {\em 8th symposium on multidisciplinary analysis and
  optimization}, page 4891, 2000.

\bibitem[\protect\citeauthoryear{Auer \bgroup \em et al.\egroup
  }{2002}]{Auer2002}
Peter Auer, Nicol{\`{o}} Cesa-Bianchi, and Paul Fischer.
\newblock {Finite-time analysis of the multiarmed bandit problem}.
\newblock {\em Machine Learning}, 47(2-3):235--256, may 2002.

\bibitem[\protect\citeauthoryear{Bai \bgroup \em et al.\egroup
  }{2016}]{bai2016markovian}
Aijun Bai, Siddharth Srivastava, and Stuart~J Russell.
\newblock Markovian state and action abstractions for {MDPs} via hierarchical
  {MCTS}.
\newblock In {\em IJCAI}, 2016.

\bibitem[\protect\citeauthoryear{Boutilier \bgroup \em et al.\egroup
  }{1999}]{Boutilier1999}
Craig Boutilier, Thomas Dean, and Steve Hanks.
\newblock Decision-theoretic planning: Structural assumptions and computational
  leverage.
\newblock {\em JAIR}, 11, 1999.

\bibitem[\protect\citeauthoryear{Browne \bgroup \em et al.\egroup
  }{2012}]{Browne2012}
Cameron~B Browne, Edward Powley, Daniel Whitehouse, Simon~M Lucas, Peter~I
  Cowling, Philipp Rohlfshagen, Stephen Tavener, Diego Perez, Spyridon
  Samothrakis, and Simon Colton.
\newblock A survey of monte carlo tree search methods.
\newblock {\em IEEE Transactions on Computational Intelligence and AI in
  games}, 2012.

\bibitem[\protect\citeauthoryear{Buesing \bgroup \em et al.\egroup
  }{2018}]{buesing2018learning}
Lars Buesing, Theophane Weber, S{\'e}bastien Racaniere, SM~Eslami, Danilo
  Rezende, David~P Reichert, Fabio Viola, Frederic Besse, Karol Gregor, Demis
  Hassabis, et~al.
\newblock Learning and querying fast generative models for reinforcement
  learning.
\newblock {\em arXiv preprint arXiv:1802.03006}, 2018.

\bibitem[\protect\citeauthoryear{Chen \bgroup \em et al.\egroup
  }{2020}]{chen2020satellite}
Qifeng Chen, Caisheng Wei, Yongfeng Shi, and Yunhe Meng.
\newblock Satellite swarm reconfiguration planning based on surrogate models.
\newblock {\em Journal of Guidance, Control, and Dynamics}, 43(9):1750--1756,
  2020.

\bibitem[\protect\citeauthoryear{Chitnis and
  Lozano-P{\'e}rez}{2020}]{chitnis2020learning}
Rohan Chitnis and Tom{\'a}s Lozano-P{\'e}rez.
\newblock Learning compact models for planning with exogenous processes.
\newblock In {\em CoRL}, 2020.

\bibitem[\protect\citeauthoryear{Cho \bgroup \em et al.\egroup
  }{2014}]{Cho2014}
Kyunghyun Cho, Bart {Van Merri{\"{e}}nboer}, Caglar Gulcehre, Dzmitry Bahdanau,
  Fethi Bougares, Holger Schwenk, and Yoshua Bengio.
\newblock {Learning phrase representations using RNN encoder-decoder for
  statistical machine translation}.
\newblock In {\em EMNLP}, 2014.

\bibitem[\protect\citeauthoryear{Congeduti \bgroup \em et al.\egroup
  }{2021}]{Congeduti21AAMAS}
Elena Congeduti, Alexander Mey, and Frans~A Oliehoek.
\newblock Loss bounds for approximate influence-based abstraction.
\newblock In {\em AAMAS}, 2021.

\bibitem[\protect\citeauthoryear{Grzeszczuk \bgroup \em et al.\egroup
  }{1998}]{grzeszczuk1998neuroanimator}
Radek Grzeszczuk, Demetri Terzopoulos, and Geoffrey Hinton.
\newblock Neuroanimator: Fast neural network emulation and control of
  physics-based models.
\newblock In {\em Proceedings of the 25th annual conference on Computer
  graphics and interactive techniques}, 1998.

\bibitem[\protect\citeauthoryear{Hansen and Feng}{2000}]{hansen2000dynamic}
Eric~A Hansen and Zhengzhu Feng.
\newblock Dynamic programming for {POMDPs} using a factored state
  representation.
\newblock In {\em AIPS}, 2000.

\bibitem[\protect\citeauthoryear{He \bgroup \em et al.\egroup
  }{2020}]{NEURIPS2020_2e6d9c60}
Jinke He, Miguel Suau~de Castro, and Frans Oliehoek.
\newblock Influence-augmented online planning for complex environments.
\newblock In {\em NeurIPS}, 2020.

\bibitem[\protect\citeauthoryear{Hostetler \bgroup \em et al.\egroup
  }{2014}]{hostetler2014state}
Jesse Hostetler, Alan Fern, and Tom Dietterich.
\newblock State aggregation in {Monte Carlo} tree search.
\newblock In {\em AAAI}, 2014.

\bibitem[\protect\citeauthoryear{Jiang \bgroup \em et al.\egroup
  }{2014}]{jiang2014improving}
Nan Jiang, Satinder Singh, and Richard Lewis.
\newblock Improving {UCT} planning via approximate homomorphisms.
\newblock In {\em AAMAS}, 2014.

\bibitem[\protect\citeauthoryear{Kaelbling \bgroup \em et al.\egroup
  }{1998}]{Kaelbling1998}
Leslie~Pack Kaelbling, Michael~L. Littman, and Anthony~R. Cassandra.
\newblock {Planning and acting in partially observable stochastic domains}.
\newblock {\em Artificial Intelligence}, 101(1-2):99--134, may 1998.

\bibitem[\protect\citeauthoryear{Kennedy and
  O'Hagan}{2000}]{kennedy2000predicting}
Marc~C Kennedy and Anthony O'Hagan.
\newblock Predicting the output from a complex computer code when fast
  approximations are available.
\newblock {\em Biometrika}, 87(1):1--13, 2000.

\bibitem[\protect\citeauthoryear{Kingma and Ba}{2014}]{kingma2014adam}
Diederik~P Kingma and Jimmy Ba.
\newblock Adam: A method for stochastic optimization.
\newblock {\em arXiv preprint arXiv:1412.6980}, 2014.

\bibitem[\protect\citeauthoryear{Oliehoek \bgroup \em et al.\egroup
  }{2021}]{Oliehoek21JAIR}
Frans Oliehoek, Stefan Witwicki, and Leslie Kaelbling.
\newblock A sufficient statistic for influence in structured multiagent
  environments.
\newblock {\em JAIR}, 70, 2021.

\bibitem[\protect\citeauthoryear{Paninski}{2003}]{paninski2003estimation}
Liam Paninski.
\newblock Estimation of entropy and mutual information.
\newblock {\em Neural computation}, 2003.

\bibitem[\protect\citeauthoryear{Peherstorfer \bgroup \em et al.\egroup
  }{2018}]{peherstorfer2018survey}
Benjamin Peherstorfer, Karen Willcox, and Max Gunzburger.
\newblock Survey of multifidelity methods in uncertainty propagation,
  inference, and optimization.
\newblock {\em Siam Review}, 60(3):550--591, 2018.

\bibitem[\protect\citeauthoryear{Pehlivanoglu and
  Yagiz}{2012}]{pehlivanoglu2012aerodynamic}
Y~Volkan Pehlivanoglu and Bedri Yagiz.
\newblock Aerodynamic design prediction using surrogate-based modeling in
  genetic algorithm architecture.
\newblock {\em Aerospace Science and Technology}, 2012.

\bibitem[\protect\citeauthoryear{Peng and Wang}{2016}]{Peng20AST}
Haijun Peng and Wei Wang.
\newblock Adaptive surrogate model-based fast path planning for spacecraft
  formation reconfiguration on libration point orbits.
\newblock {\em Aerospace Science and Technology}, 54:151--163, 2016.

\bibitem[\protect\citeauthoryear{Ruder}{2016}]{Ruder2016}
Sebastian Ruder.
\newblock {An overview of gradient descent optimization algorithms}.
\newblock {\em arXiv preprint arXiv:1609.04747}, sep 2016.

\bibitem[\protect\citeauthoryear{Silver and Veness}{2010}]{Silver2010}
David Silver and Joel Veness.
\newblock {Monte-Carlo planning in large POMDPs}.
\newblock In {\em NeurIPS}, 2010.

\bibitem[\protect\citeauthoryear{Somani \bgroup \em et al.\egroup
  }{2013}]{NIPS2013_c2aee861}
Adhiraj Somani, Nan Ye, David Hsu, and Wee~Sun Lee.
\newblock Despot: Online pomdp planning with regularization.
\newblock In {\em NeurIPS}, 2013.

\bibitem[\protect\citeauthoryear{Spaan}{2012}]{spaan2012partially}
Matthijs~TJ Spaan.
\newblock Partially observable markov decision processes.
\newblock In {\em Reinforcement Learning}, pages 387--414. Springer, 2012.

\end{thebibliography}

\clearpage

\appendix

{\onecolumn

\section{Full derivation for the estimation of accuracy measure}\label{appendix:full-derivation}

In the following, we provide full derivation for the results presented in the subsection Estimating the Accuracy. 

\subsection{Cross entropy estimation}\label{appendix:full-derivation-cross-entropy}
\begin{align*}
  \E_{d_k \sim P_k(\cdot | h_t, \pi^{\tree_i})} H(I(\Ssrc_k | d_k), \aip(\Ssrc_k | d_k)) &= - \sum_{d_k} P_k(d_k | h_t, \pi^{\tree_i}) \sum_{s^{src}_k} I(s^{src}_k | d_k) \log{\aip(s^{src}_k | d_k))} \\
  &= - \sum_{d_k, s^{src}_k} P(d_k, s^{src}_k | h_t, \pi^{\tree_i}) \log{\aip(s^{src}_k | d_k)} \\
  &= - \sum_{d_k, s_k} P(d_k, s_k=<\ssrc_k, s_k /\ \ssrc_k> | h_t, \pi^{\tree_i}) \log{\aip(s^{src}_k | d_k)} \\
  &\text{\{$s_k$ is the global state that contains all the state variables, including $\ssrc_k$\}} \\
  &= - \sum_{d_k, s_k} P(d_k, s_k | h_t, \pi^{\tree_i}) \log{\aip(s^{src}_k | d_k)} \\ 
  &= - \E_{d_k, s_k \sim P(\cdot | h_t, \pi^{\tree_i})} \log{\aip(s^{src}_k | d_k)} 
\end{align*}

\subsection{Entropy estimation}\label{appendix:full-derivation-entropy}
\begin{align*}
  \E_{d_k \sim P_k(\cdot | h_t, \pitree)} H(I(\Ssrc_k | d_k)) &= H(S^{src}_k | D_k, h_t, \pitree) \\
  &\geq \; H(S^{src}_k | S_{k-1}, A_{k-1}, D_k, h_t, \pitree)) \\
  &\text{\{since adding information does not increase entropy\}} \\ 
  &= \sum_{s_{k-1}, a_{k-1}, d_k} P{(s_{k-1}, a_{k-1}, d_k | h_t, \pitree)} H(S^{src}_{k}|s_{k-1}, a_{k-1}, d_{k}=(d_{k-1}, a_{k-1}, s^{local}_k)) \\  
  &= \sum_{s_{k-1}, a_{k-1}, d_k} P(s_{k-1}, a_{k-1}, d_k | h_t, \pitree)  H(S^{src}_k | s_{k-1}, a_{k-1}, s^{local}_k) \\
  &\text{\{$d_{k-1}=\{\slocal_0, a_0, \slocal_1, \ldots, a_{k-2}, \slocal_{k-1}\}$ adds no extra information due to markov property\}} \\ 
  &= \sum_{s_{k-1}, a_{k-1}, d_k} P(s_{k-1}, a_{k-1}, d_k | h_t, \pitree) H(S^{src}_k | s_{k-1}, a_{k-1})  \\
  &\text{\{here we use the assumption there is no inter-stage dependency between state variables\}} \\
  &= \sum_{s_{k-1}, a_{k-1}} P(s_{k-1}, a_{k-1}| h_t, \pitree) H(S^{src}_k | s_{k-1}, a_{k-1} ) \\
  &= \E_{s_{k-1}, a_{k-1} \sim P(\cdot | h_t, \pitree) } H(S^{src}_k | s_{k-1}, a_{k-1}) 
\end{align*}

\section{Pseudocode for the two-phase approach by [He \emph{et al.}, 2020]}


\begin{algorithm}[h]
\small
\caption{Planning with IALS [He \emph{et al.}, 2020]}
\label{alg:planIALS}
\begin{algorithmic}[1] 
\renewcommand{\algorithmicrequire}{\textbf{Offline Training Phase:}}
\renewcommand{\algorithmicensure}{\textbf{Online Planning Phase:}}
\REQUIRE 
\STATE collect training data $\D$ with policy $\piexplore$ from $\simGM$
\STATE train the influence predictor $\aip$ on $\D$ to convergence
\ENSURE
\FOR{every real episode} 
\FOR{every time step $t$} 
\FOR{every simulation $i=0,\ldots$} 
\STATE simulate a trajectory $\tau_i$ with $\simIALMp$
\STATE update the search tree with $\tau_i$
\ENDFOR
\STATE take the greedy action and prune the tree with new observation
\ENDFOR
\ENDFOR
\end{algorithmic}
\end{algorithm}

\clearpage

\section{Details of the experimental setup}\label{appendix:experimental-setup}

In the following, we describe the further details of the experimental setup, including the planning domains and the hyperparameters that are used.

\subsection{Domains}\label{appendix:domains}

\begin{figure*}[t]
    \centering
    \begin{subfigure}[b]{0.49\textwidth}
        \centering
        \includegraphics[width=.6\textwidth]{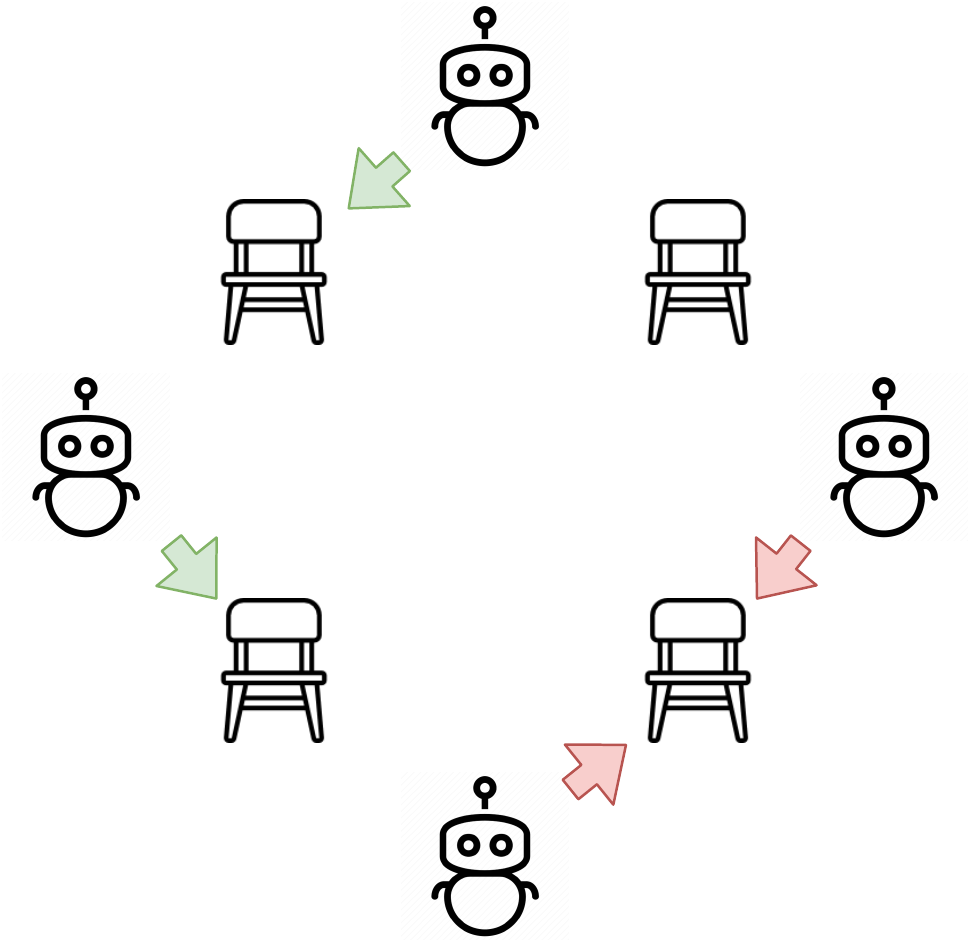}
        \caption{}
        \label{env:GAC}
    \end{subfigure}
    \begin{subfigure}[b]{0.49\textwidth}
        \centering
        \includegraphics[width=.6\textwidth]{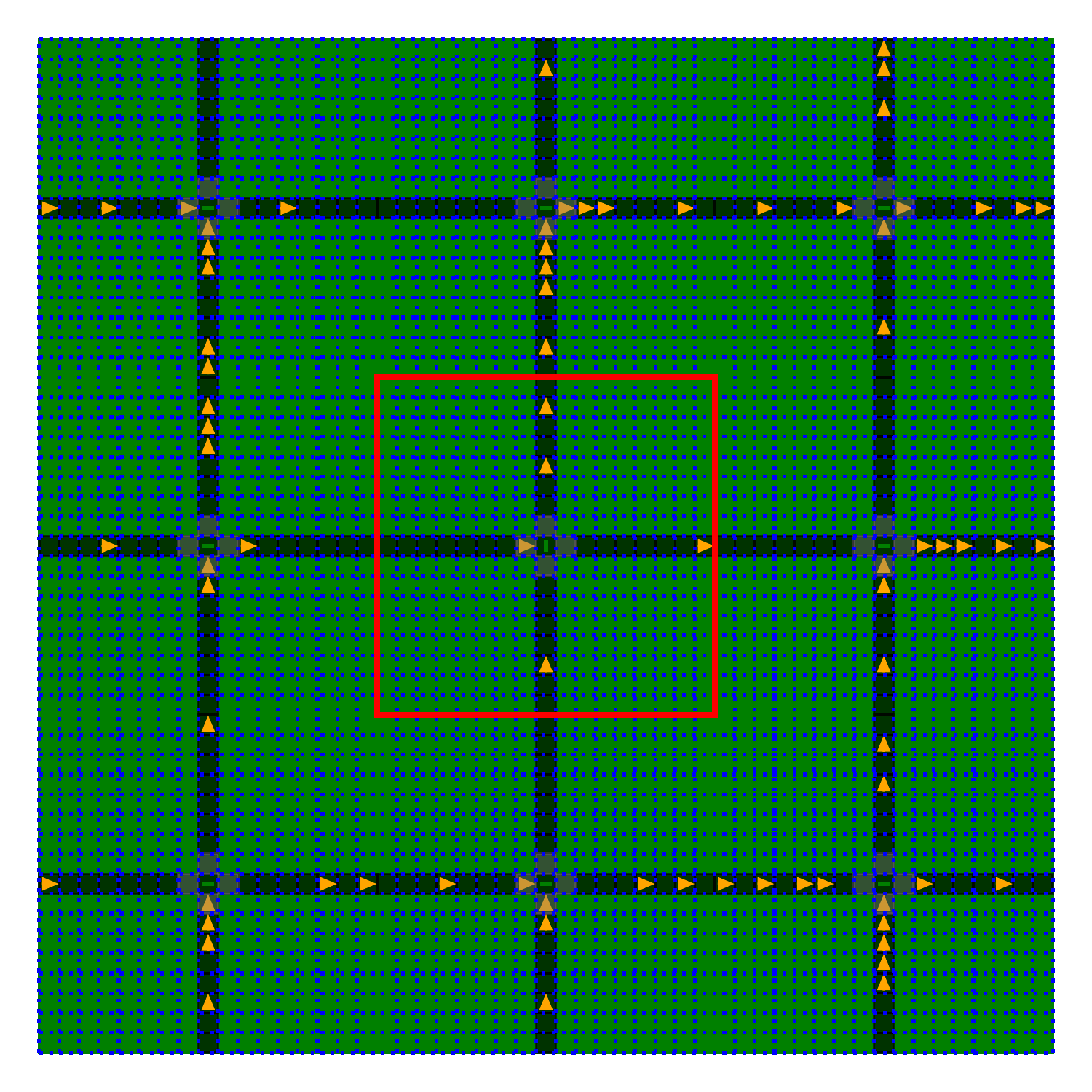}
        \caption{}
        \label{env:GTC}
    \end{subfigure}
    \caption{(a) An example of the grab a chair domain with $4$ agents. For this time step, the agents on the top and left successfully obtain a chair and receive a reward $1$ while the other two do not because they target the same chair. (b) A visualization of the grid traffic control domain with yellow arrows represent the moving cars. The intersection controlled by the agent is highlighted by a red bounding box.}
    \label{fig:env}
\end{figure*}

\paragraph{Grab a chair domain} In our experiments, we make use of a grab a chair domain that lasts for $10$ steps with one planning agent and $64$ other agents executing fixed policies. At time step $t$, each agent has two possible actions which are grabbing the chair on its left and right side. When all the decisions are made, every agent which makes the decision to grab a chair that is not targeted by another neighboring agent will get the chair, and receive a reward of $1$, otherwise none of the agents get the chair and both of them will receive $0$ reward. The agent can only observe whether itself has obtained the chair or not, with a noise rate of $0.2$. The fixed policy that is used by the $64$ fixed agents make decisions based on the empirical frequency of obtaining a chair on the left and right side, according to their action-observation histories so far. See Figure \ref{env:GAC} for an example.

\paragraph{Grid traffic control domain} As illustrated by Figure \ref{env:GTC}, the grid traffic control domain is made of $9$ traffic intersections in total where the planning agent controls the one in the middle. At every time step $t$, the agent can take an action to switch the traffic light of the local intersection that it controls. The agent has access to a sensor data on if there are cars at the four grids that surround the traffic light. The reward of the agent is the negative total number of cars in this intersection at this time step. For the other $8$ intersections in this intersection, they use a policy that switches their traffic lights every $9$ time steps. New cars will enter the system with a probability $0.7$ and cars at the borders of the system will leave with a probability $0.3$. Initially, cars are generated with a probability $0.7$ in every grid. The horizon of this domain is $50$. 

\subsection{Hyperparameters}

\subsubsection{Planning with self-improving simulators}

The following hyperparameters are used for the experiments that correspond to Figure~\ref{fig:simulation-controlled-results} (a-c) and Figure~\ref{fig:real-time-results}.

\paragraph{Influence Predictor}
In all experiments, the approximate influence predictor $\aip$ is parameterized by a gated recurrent unit (GRU) \cite{Cho2014}, a variant of recurrent neural networks, with $8$ hidden states. We use Adam \cite{kingma2014adam} as the optimizer for training with stochastic gradient descent. 

\paragraph{Training}
The online training of the influence predictors $\aip$ occurs after every real episode. During every training process, we perform $64$ steps of stochastic gradient descent with the Adam optimizer. The data is sampled from the replay buffer that stores all the data collected so far, with a batch size of $128$. The learning rates we use for the grab a chair and grid traffic control domains are $0.001$ and $0.00025$, respectively.

\paragraph{Planning}

In Table \ref{table:hyper-GAC} and Table \ref{table:hyper-GTC} we list the hyperparameters for planning with self-improving simulators in the grab a chair and grid traffic control domains. 

\begin{table}[H]
  \centering
  \begin{tabular}{ |c|c| } 
   \hline
   Discount factor $\gamma$ & $1.0$  \\ 
   Number of initial particles & $1000$ \\
   Exploration constant in the UCB1 algorithm $c$ & $100.0$  \\ 
   Meta exploration constant for simulation selection $c^{\mathtt{meta}}$ & $0.3$ \\
   \hline
  \end{tabular}
    \caption{Hyperparameters for planning with self-improving simulators in the grab a chair domain.}
  \label{table:hyper-GAC}
\end{table}

\begin{table}[H]
  \centering
  \begin{tabular}{ |c|c| } 
   \hline
   Discount factor $\gamma$ & $0.95$  \\ 
   Effective Horizon & $36$  \\
   Number of initial particles & $1000$ \\
   Exploration constant in the UCB1 algorithm $c$ & $10.0$  \\ 
   Meta exploration constant for simulation selection $c^{\mathtt{meta}}$ & $0.1$ \\
   \hline
  \end{tabular}
    \caption{Hyperparameters for planning with self-improving simulators in the grid traffic control domain.}
  \label{table:hyper-GTC}
\end{table}



\section{Additional results}\label{appendix:additional-results}

\subsection{The simulation controlled planning experiments}

In Figure \ref{fig:add-fixed_sim} we provide the additional results for the planning experiments with fixed number of simulations per step in the grab a chair domain. The left figure shows the learning curves of the influence predictor over real episodes. The right figure shows the estimated inaccuracy for the IALS. As we can see, due to training, generally the estimated inaccuracy is decreasing, which is expected. This leads to the increasing use of the IALS as shown in the main text, thus speeding up planning. We can also see that with $\lambda=3.0$ and $\lambda=1.5$, the estimated inaccuracy is much lower in the beginning than the other $\lambda$s. This is because with a large $\lambda$, the accuracy  threshold to use the IALS is much lower, which results that the untrained IALSs are used exclusively right from start. This limits the number of global simulations, causing poor accuracy of the inaccuracy estimation itself. 

\subsection{The (time controlled) real-time planning experiments}

In Figure~\ref{fig:add-real-time-GAC} and \ref{fig:add-real-time-GTC}, we provide the additional results for the real-time planning experiments.  The common trend is that, over real episodes, the training loss of the approximate influence predictor $\aip$ (used in the IALS) decreases, which is translated into the decreasing inaccuracy estimate. This leads to more planning time spent on the IALS, which results in more simulations being done within the same time limit, since it is significantly faster than the global simulator. 

\begin{figure*}[t]
  \centering
  \begin{subfigure}[b]{0.45\textwidth}
      \centering
      \includegraphics[scale=0.42]{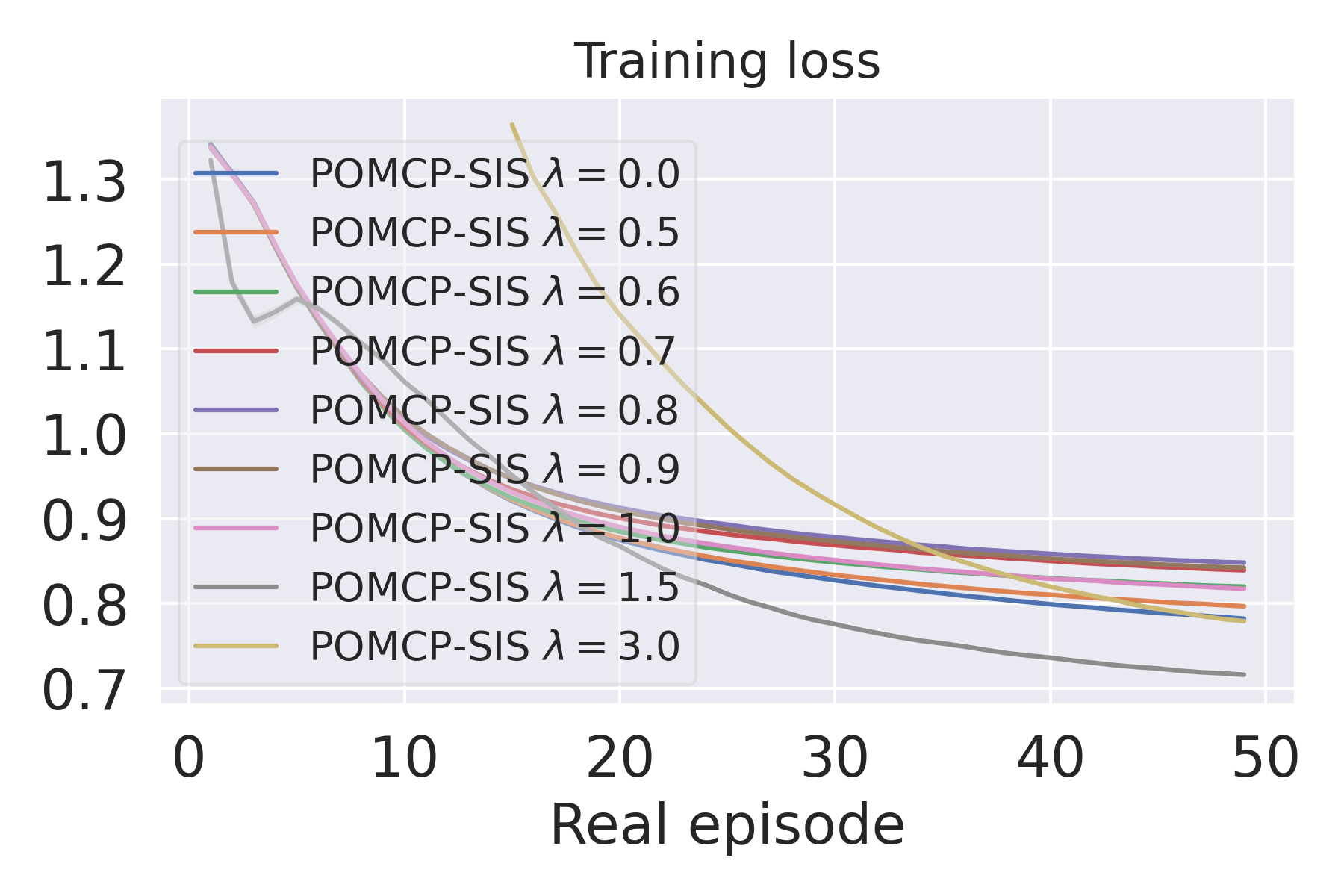}
      \caption{}
      \label{fig:fixed-sim-loss}
  \end{subfigure}
  \begin{subfigure}[b]{0.45\textwidth}
      \centering
      \includegraphics[scale=0.42]{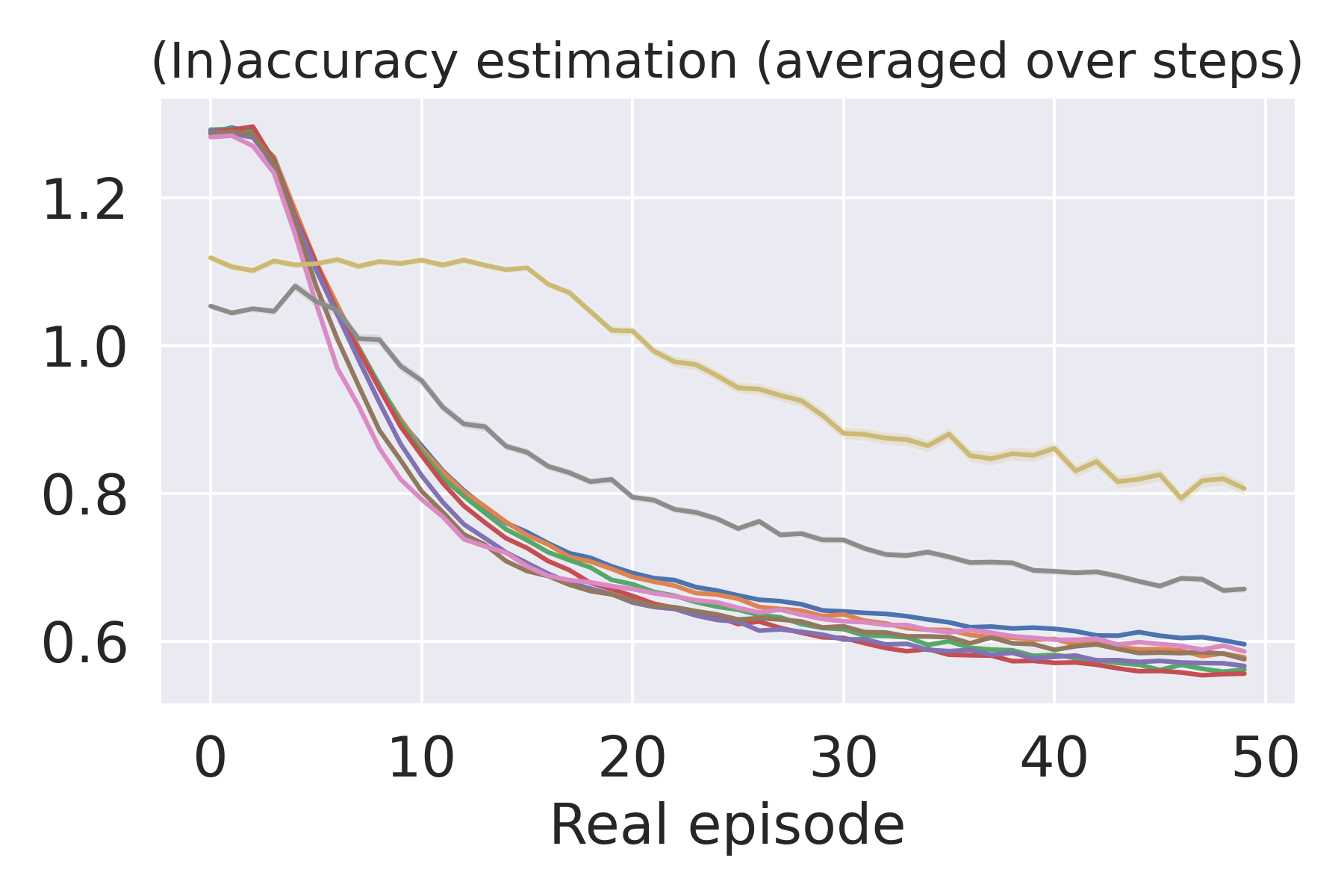}
      \caption{}
      \label{fig:fixed-sim-kl}
  \end{subfigure}
  \caption{Additional results for the simulation controlled planning experiments in the grab a chair domain (accompanying Figure~\ref{fig:simulation-controlled-results} (a-c)).}
  \label{fig:add-fixed_sim}
\end{figure*}

\begin{figure*}[t]
  \centering
  \begin{subfigure}[b]{0.45\textwidth}
      \centering
      \includegraphics[scale=0.42]{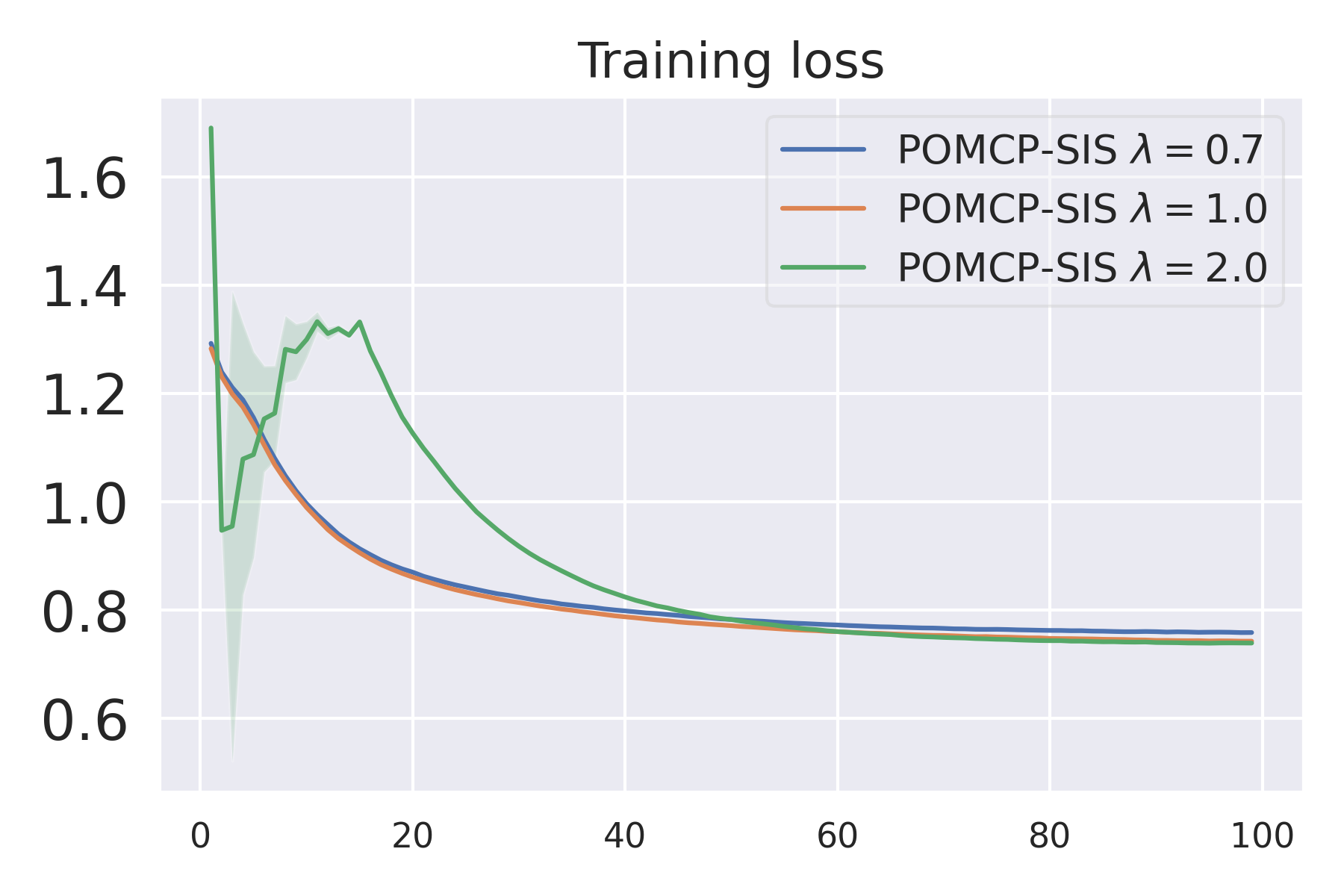}
      \caption{}
      \label{fig:real-time-GAC-loss}
  \end{subfigure}
  \begin{subfigure}[b]{0.45\textwidth}
      \centering
      \includegraphics[scale=0.42]{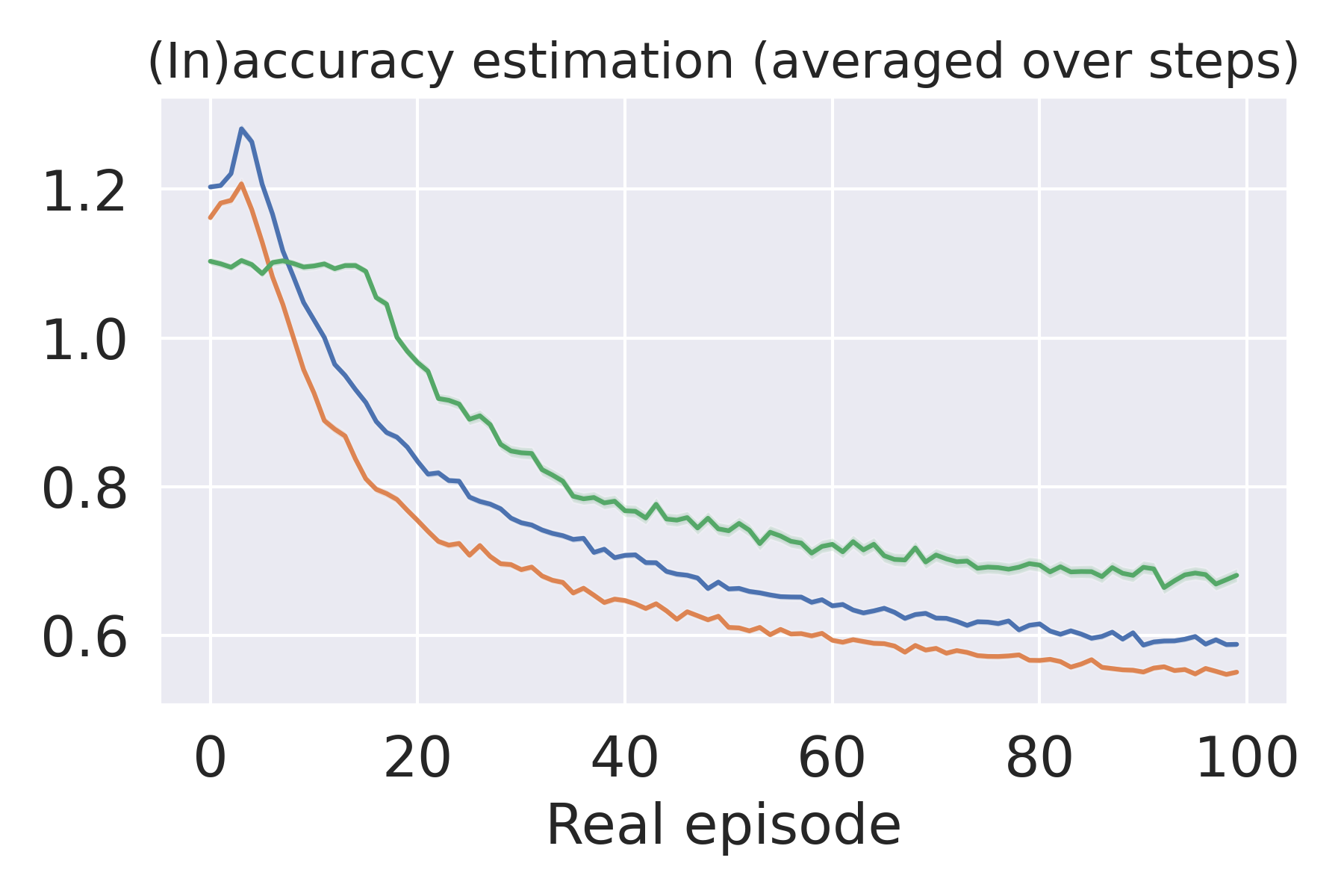}
      \caption{}
      \label{fig:real-time-GAC-kl}
  \end{subfigure}
  \begin{subfigure}[b]{0.45\textwidth}
    \centering
    \includegraphics[scale=0.42]{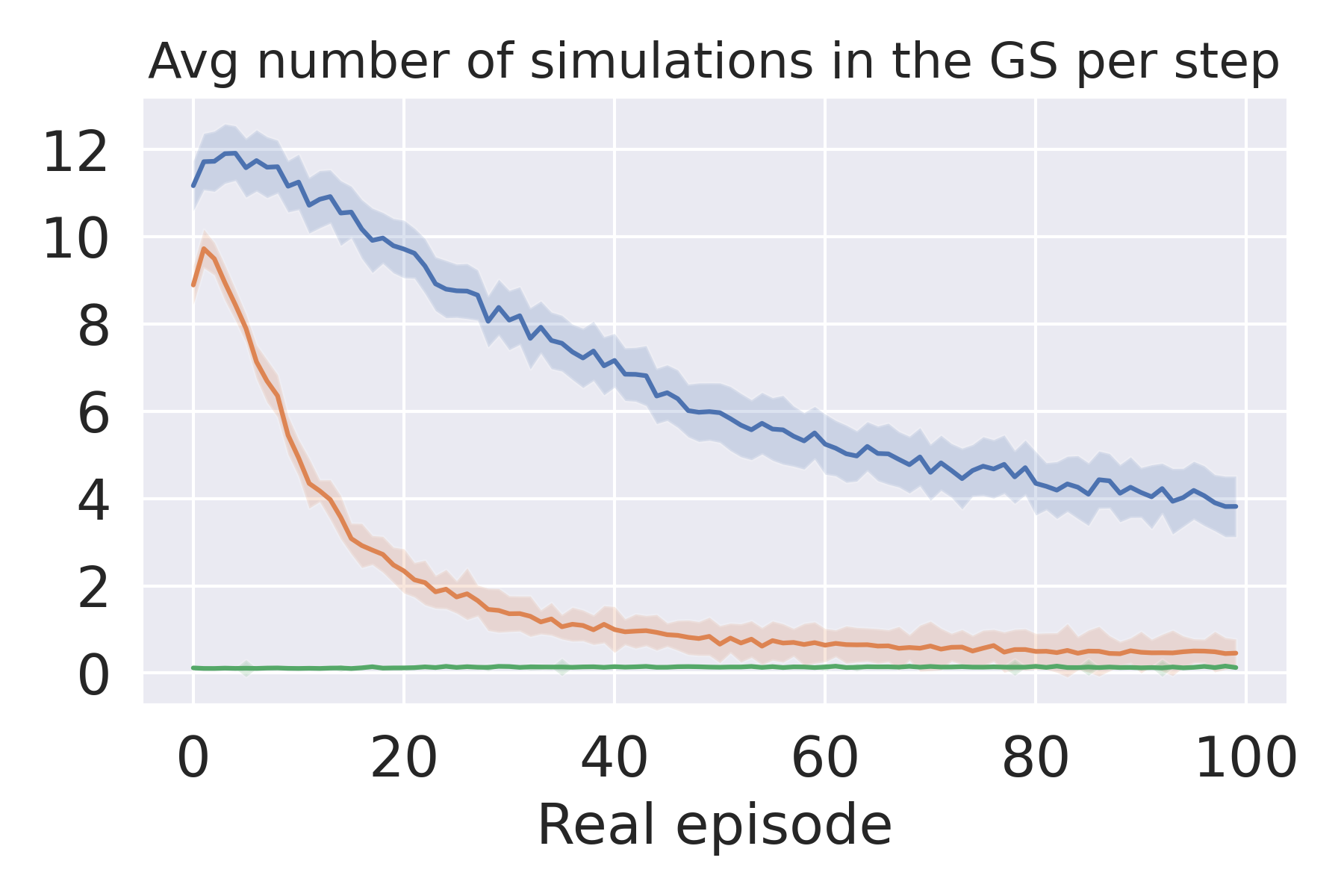}
    \caption{}
    \label{fig:real-time-GAC-gs-counts}
  \end{subfigure}
  \begin{subfigure}[b]{0.45\textwidth}
    \centering
    \includegraphics[scale=0.42]{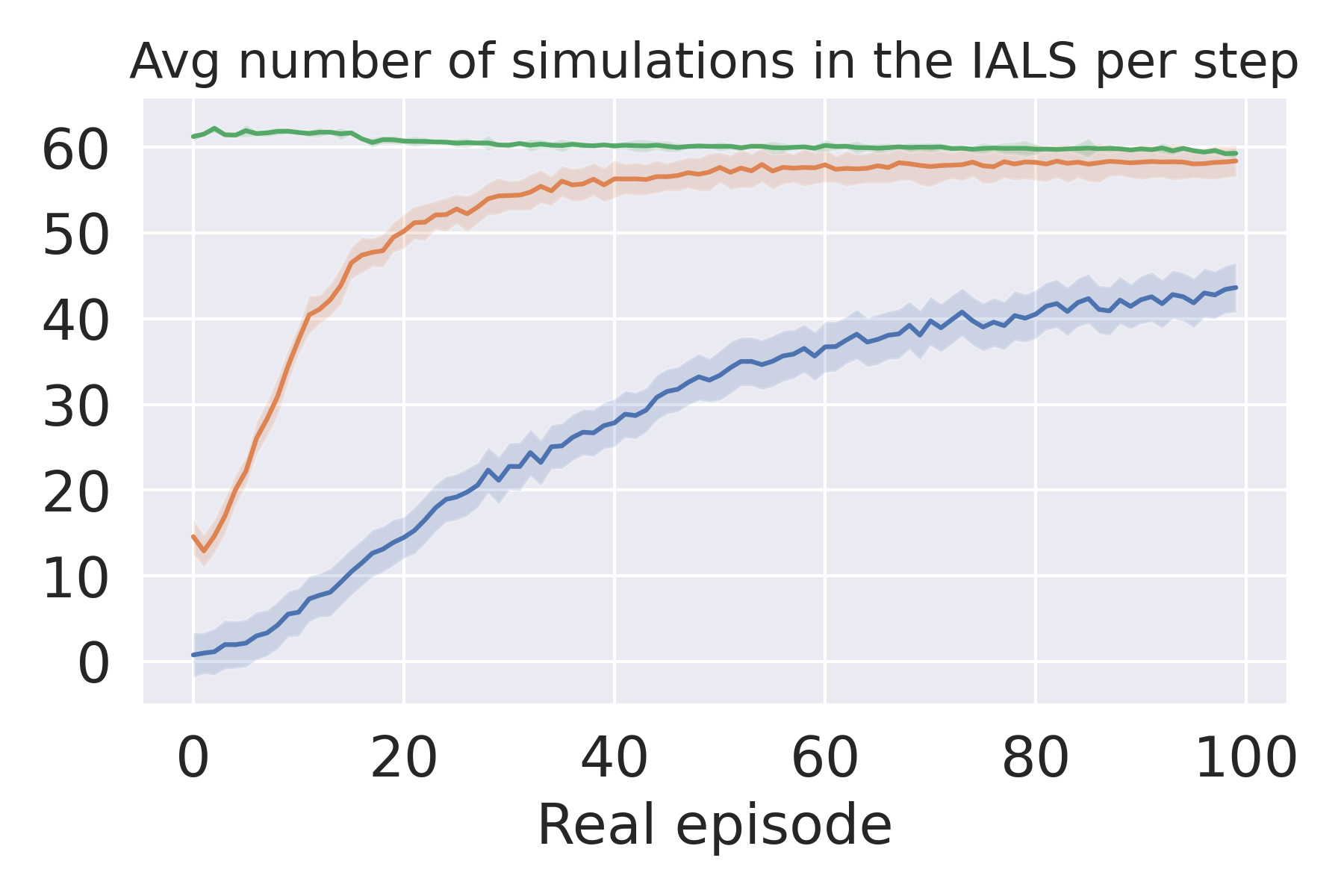}
    \caption{}
    \label{fig:real-time-GAC-ials-counts}
  \end{subfigure}
  \caption{Additional results for the real-time planning experiments in the grab a chair domain (accompanying Figure~\ref{fig:real-time-GAC}).}
  \label{fig:add-real-time-GAC}
\end{figure*}

\begin{figure*}[t]
  \centering
  \begin{subfigure}[b]{0.45\textwidth}
      \centering
      \includegraphics[scale=0.42]{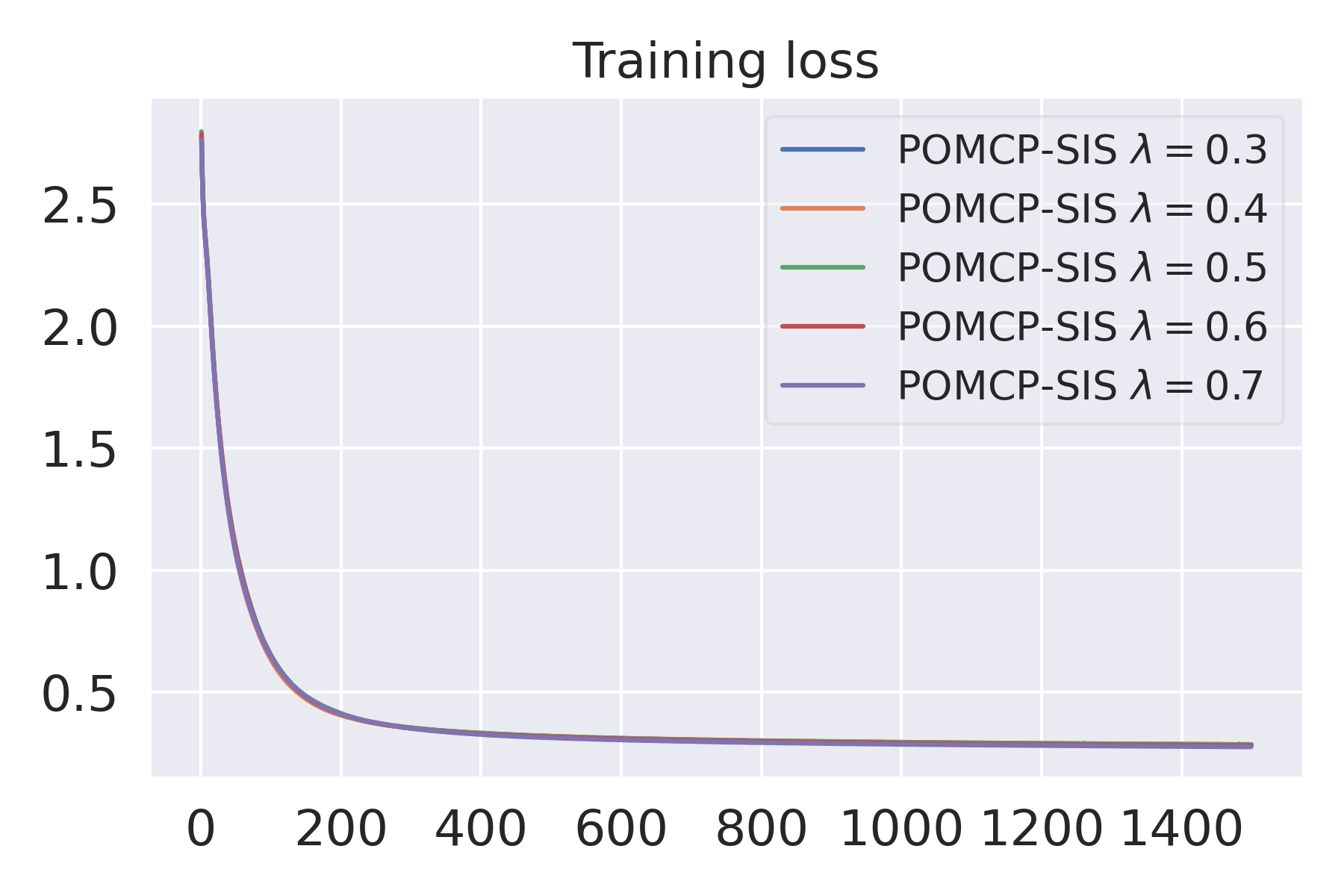}
      \caption{}
      \label{fig:real-time-GTC-loss}
  \end{subfigure}
  \begin{subfigure}[b]{0.45\textwidth}
      \centering
      \includegraphics[scale=0.42]{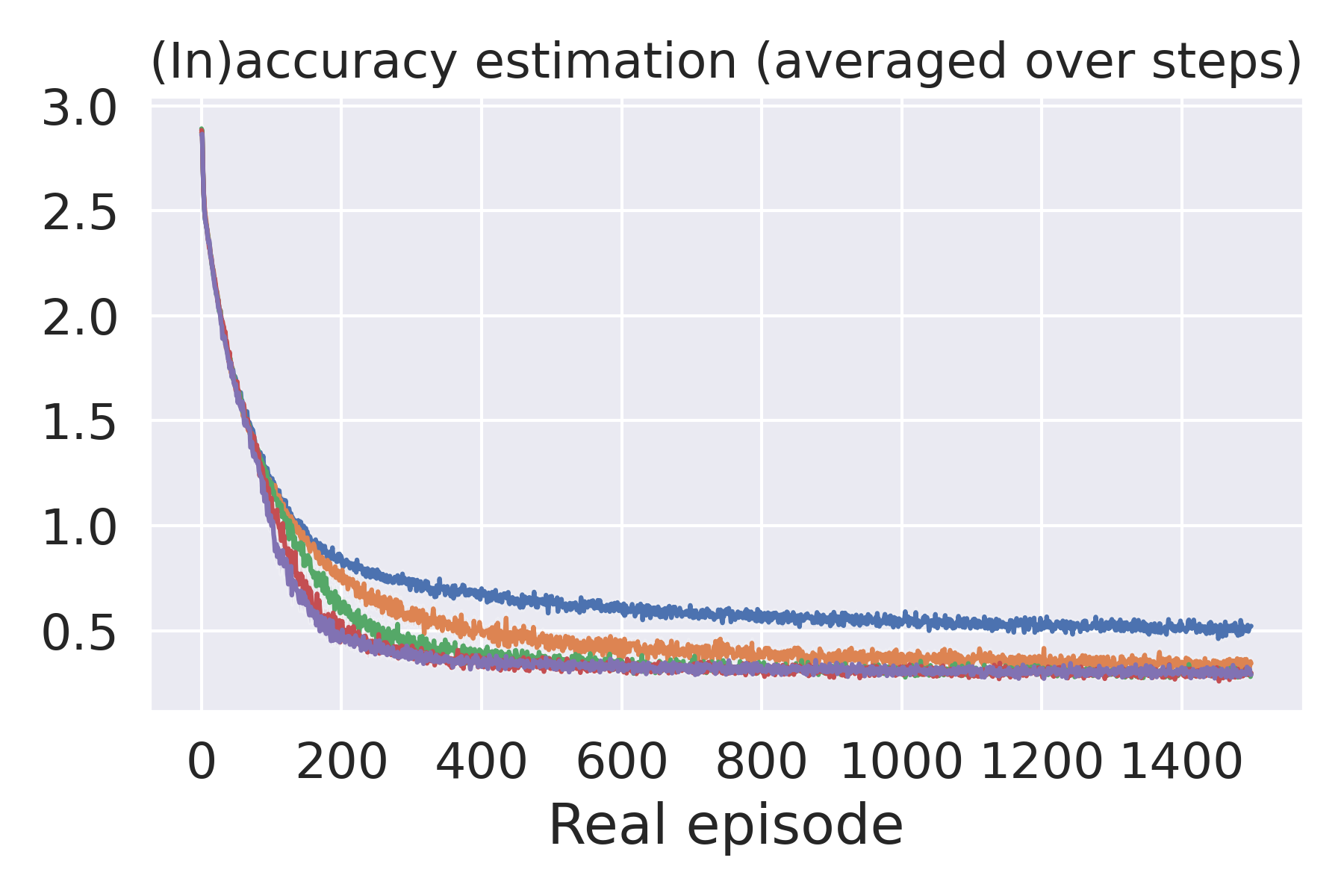}
      \caption{}
      \label{fig:real-time-GTC-kl}
  \end{subfigure}
  \begin{subfigure}[b]{0.45\textwidth}
    \centering
    \includegraphics[scale=0.42]{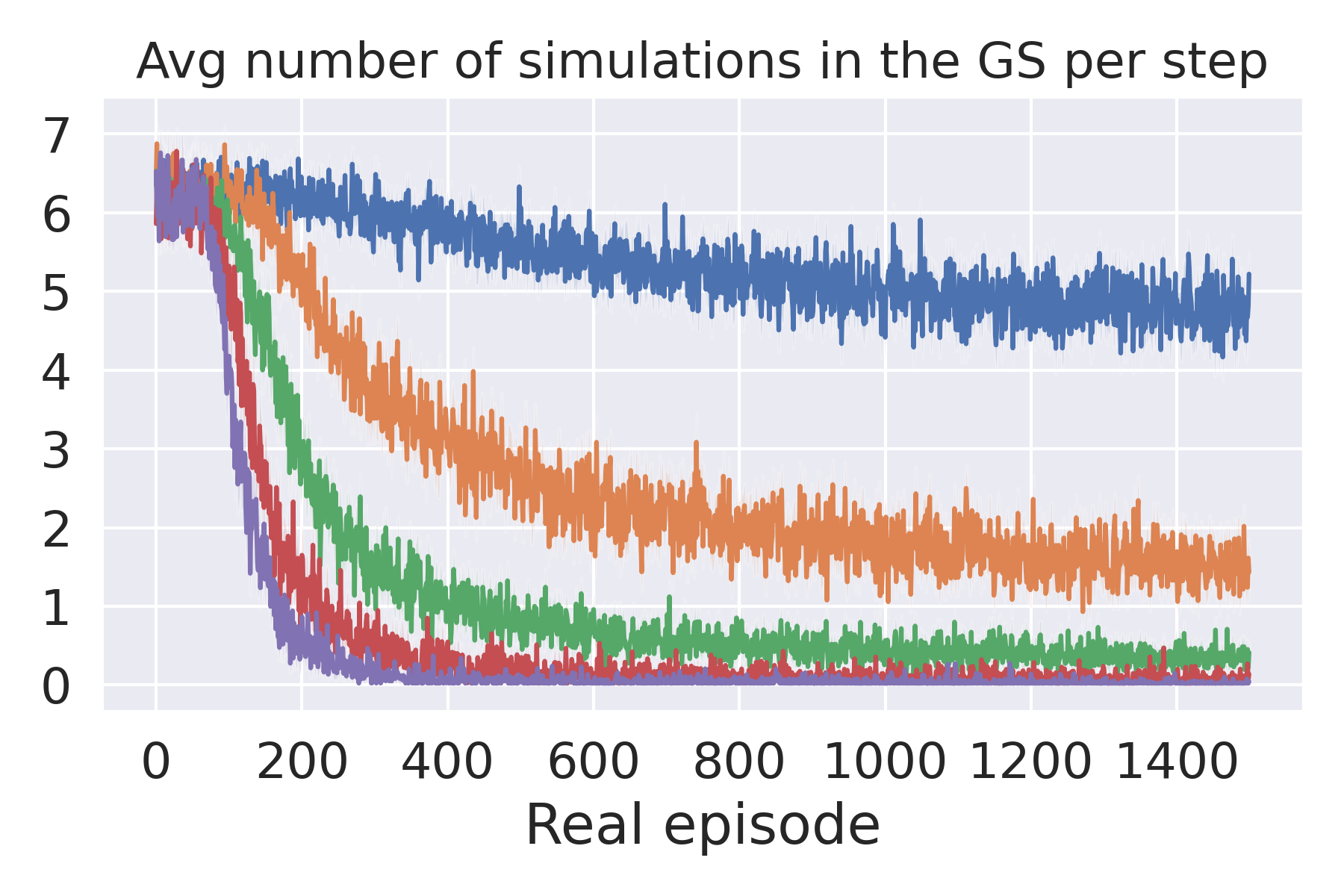}
    \caption{}
    \label{fig:real-time-GTC-gs-counts}
  \end{subfigure}
  \begin{subfigure}[b]{0.45\textwidth}
    \centering
    \includegraphics[scale=0.42]{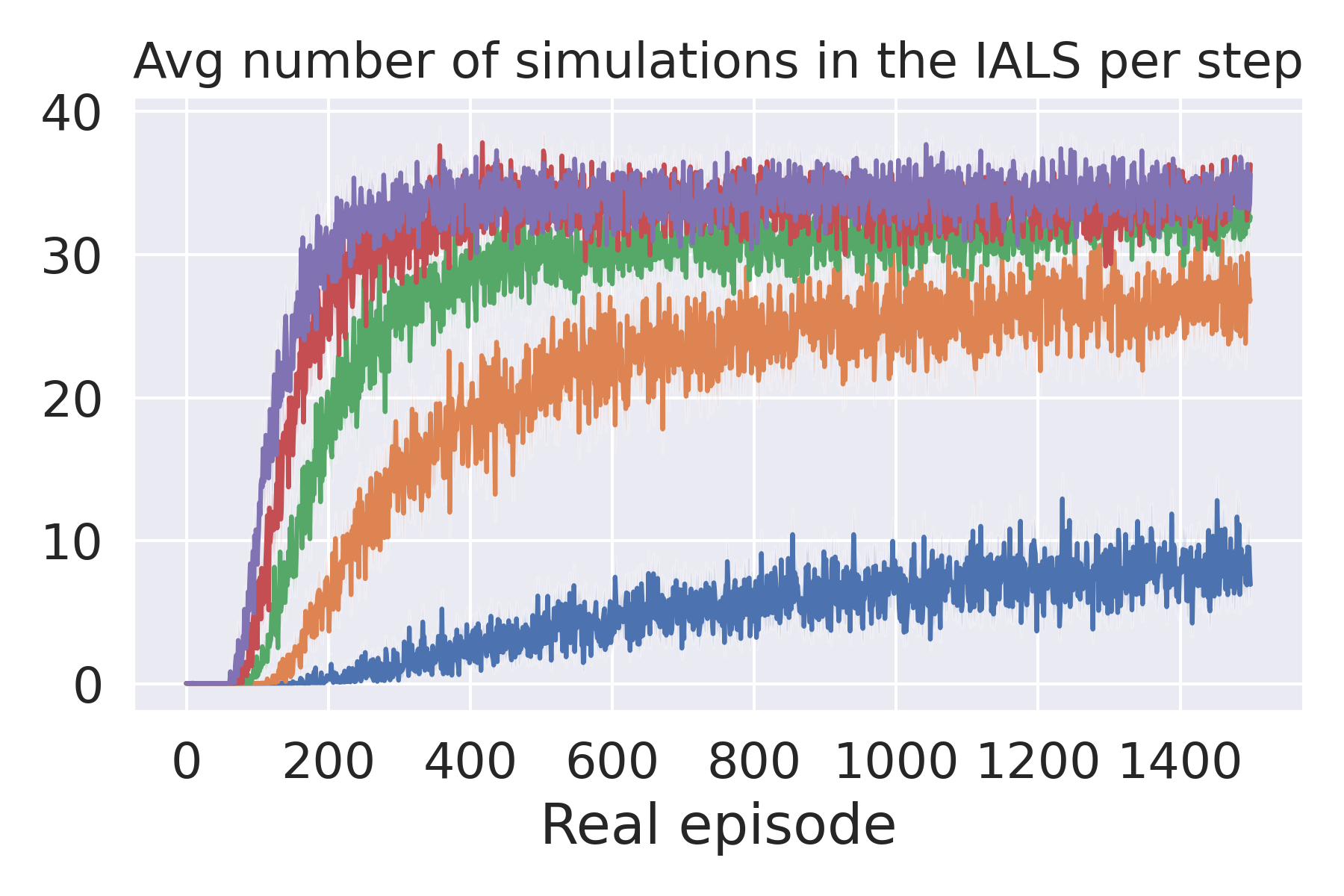}
    \caption{}
    \label{fig:real-time-GTC-ials-counts}
  \end{subfigure}
  \caption{Additional results for the real-time planning experiments in the grid traffic control domain (accompanying Figure~\ref{fig:real-time-GTC}).}
  \label{fig:add-real-time-GTC}
\end{figure*}

\subsection{The comparison to the two-phase approach}

\begin{figure}[t]
    \centering
    \includegraphics[width=0.6\textwidth]{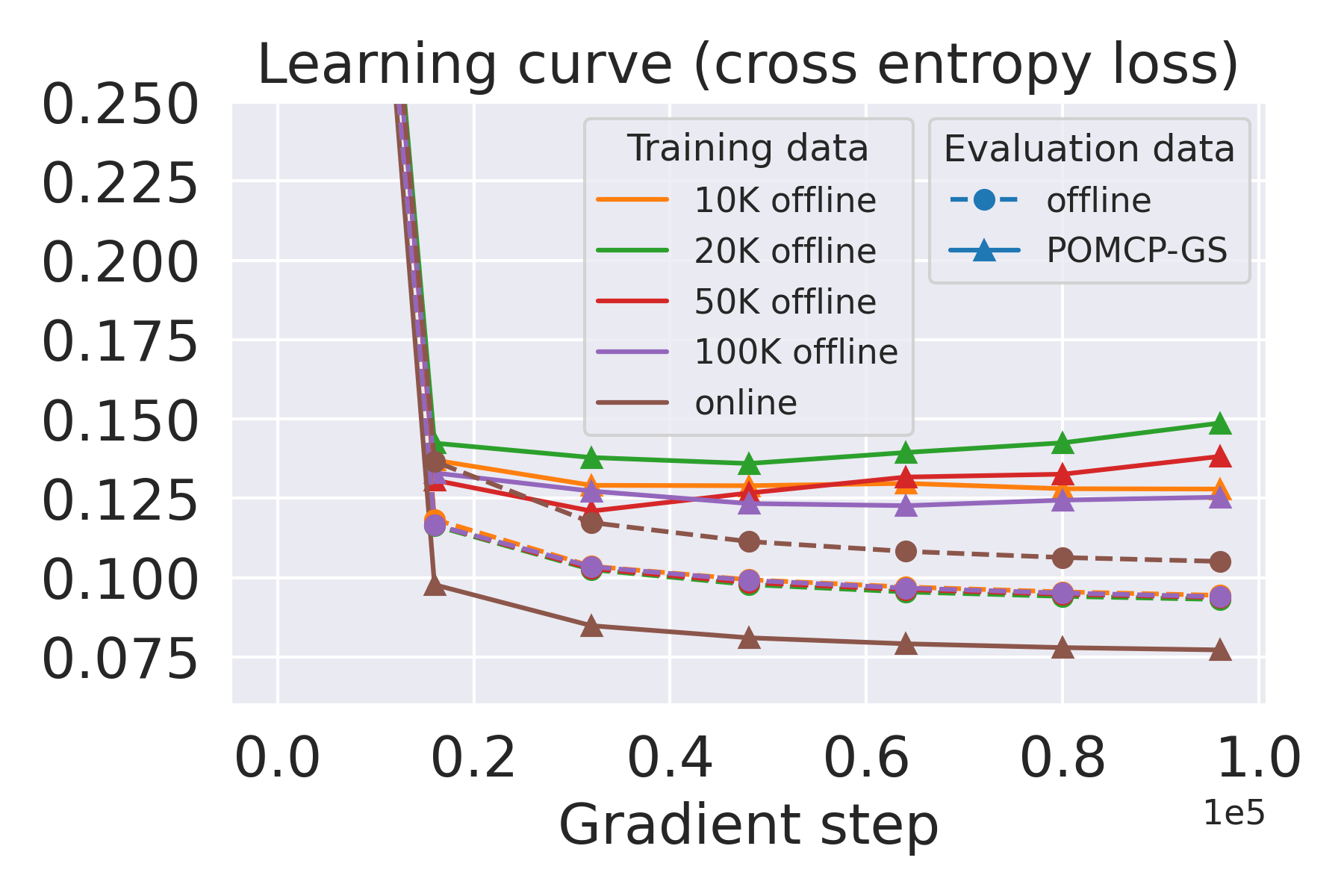}
    \caption{Learning curves of the influence predictors that are trained in an offline manner with the offline and online data (accompanying Figure~\ref{fig:offline-comparison-return}).}
    \label{fig:offline-comparison-loss}
\end{figure}

Figure~\ref{fig:offline-comparison-return} shows that the planning performance is dramatically different when training the influence predictors with the offline and online data. Note that in both cases, the influence predictors are trained offline and then used as-is for online planning. In the following we investigate the issue behind the failure of the influence predictors that are trained with data collected by a uniform random policy. To evaluate the influence predictors, we collect two independent datasets. One contains again the offline data that is collected by a uniform random policy, which should have the same distribution with the offline training data. To understand the behaviour of the influence predictors when they make online predictions during planning, we execute POMCP with the global simulator to collect the second dataset. We consider the POMCP-GS dataset as the test set here because that represents the distribution that we may encounter when planning with an exact simulator. We plot the learning curves of the influence predictors that are trained on $\{10K, 20K, 50K, 100K\}$ episodes of offline data and on the "online" data that is collected when planning with the self-improving simulator $\lambda=0.7$, importantly, evaluated on both datasets. The dotted lines with circles represent the evaluations with the offline dataset and the solid lines with triangles represent the evaluations with the POMCP-GS dataset. We can see that first of all, for all the influence predictors that are trained with offline data, there is a trend that as training goes on, the training error (evaluated on the offline dataset) is decreasing while the test error (on the POMCP-GS) dataset is increasing, a classic indicator of overfitting. This is strong evidence that there is indeed a distribution shift when training the influence predictors with data collected offline. Moreover, we see that this is not happening to the influence predictors that are trained with online data from the self-improving simulators. In the end, they can converge to a much lower test error, which can explain their much better planning performance. As such, we can conclude that this experiment demonstrates the distribution shift issue of the two-phase approach by \cite{NEURIPS2020_2e6d9c60}, and shows that planning with self-improving simulators can fix the issue. 

\subsection{Ablation study: the effect of the meta exploration constant}

To understand the effect of the meta exploration constant $c^{\mathtt{meta}}$, we repeat the simulation controlled experiment with a set of different values $c^{\mathtt{meta}}=\{0.0, 0.3, 1.0, 2.0\}$. The results are shown in Figure~\ref{fig:ablation-mec}. In the main text, we describe that there is an exploration and exploitation problem when selecting the simulators online, which we address with the UCB1 algorithm by formulating it as a bandit problem. Here we investigate the effect of the meta exploration constant on the planning performance. As we can see from Figure~\ref{fig:ablation-mec}, with a meta exploration constant $c^{\mathtt{meta}}=0.0$, which effectively removes the UCB1 action selection, there is a period at the beginning of planning, during which the agent does not perform well, for many values of $\lambda$s. The number of IALS simulations suggests this is due to already using the IALS a lot while it has not been trained much. However, this does not happen to the other values of the meta exploration constant. Our understanding is that this is due to the poor estimation of the IALS inaccuracy, i.e., a lack of exploration. On the other hand, from Figure~\ref{fig:ablation-mec}, we can see that the use of a large meta exploration constant $c^{\mathtt{meta}}=2.0$ results in a limited exploitation of the faster IALS, even when it is sufficiently trained, preventing the planning speed to increase further after reaching certain level. 

\begin{figure*}[t]
    \centering
    \begin{subfigure}[b]{\textwidth}
        \includegraphics[width=0.32\textwidth]{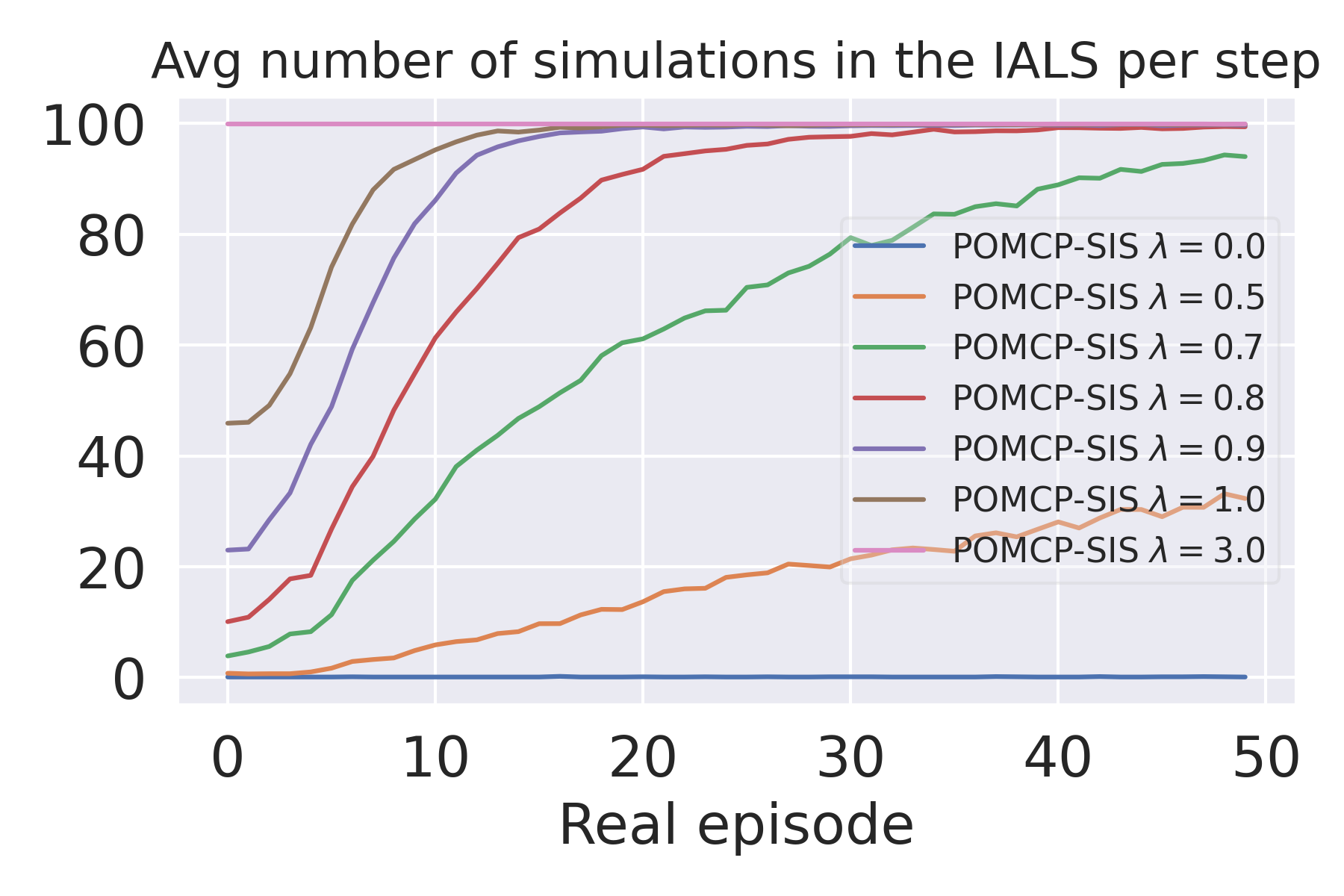}
        \includegraphics[width=0.32\textwidth]{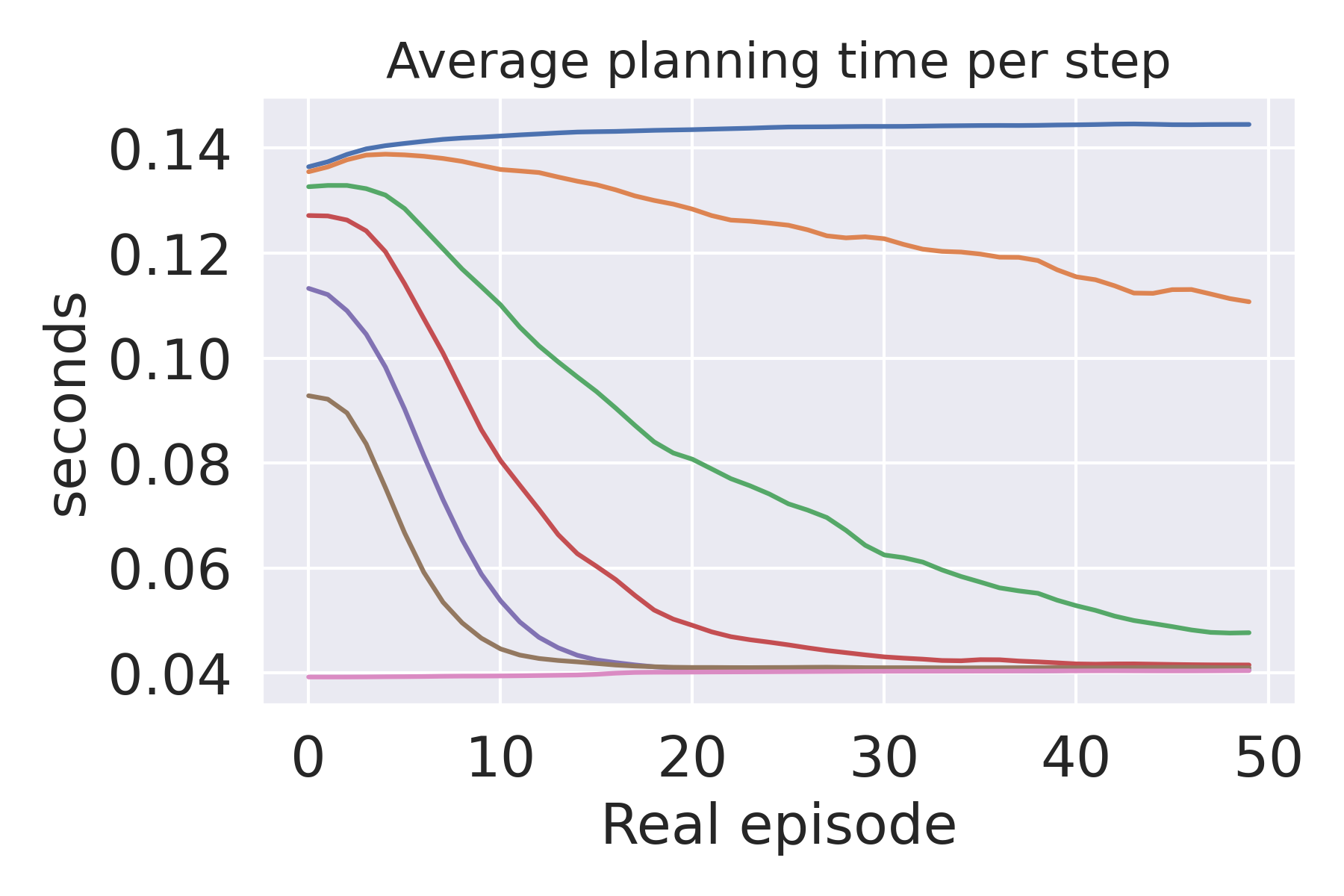}
        \includegraphics[width=0.32\textwidth]{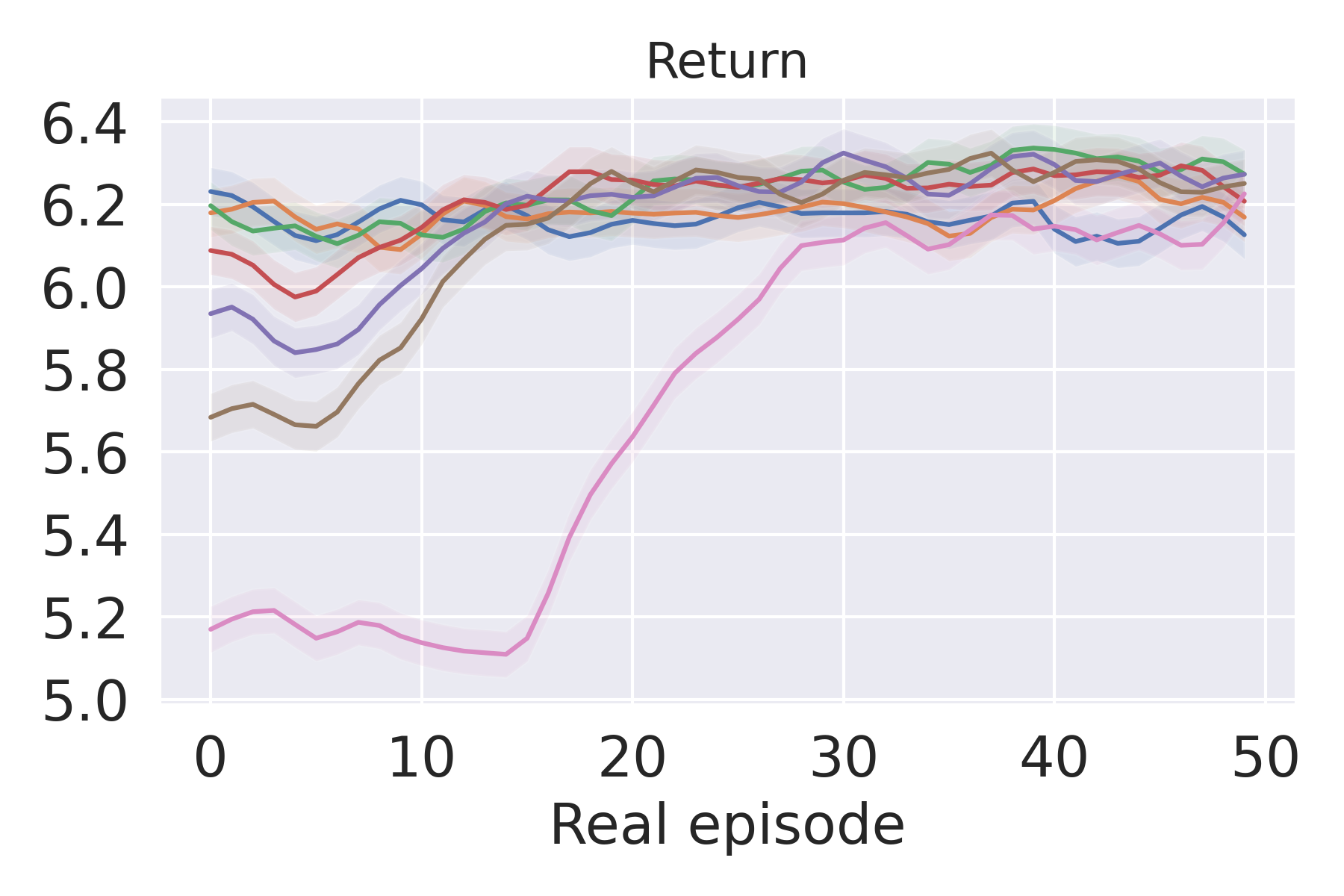}
        \caption{$c^{\mathtt{meta}}=0.0$}
    \end{subfigure}
    \begin{subfigure}[b]{\textwidth}
        \includegraphics[width=0.32\textwidth]{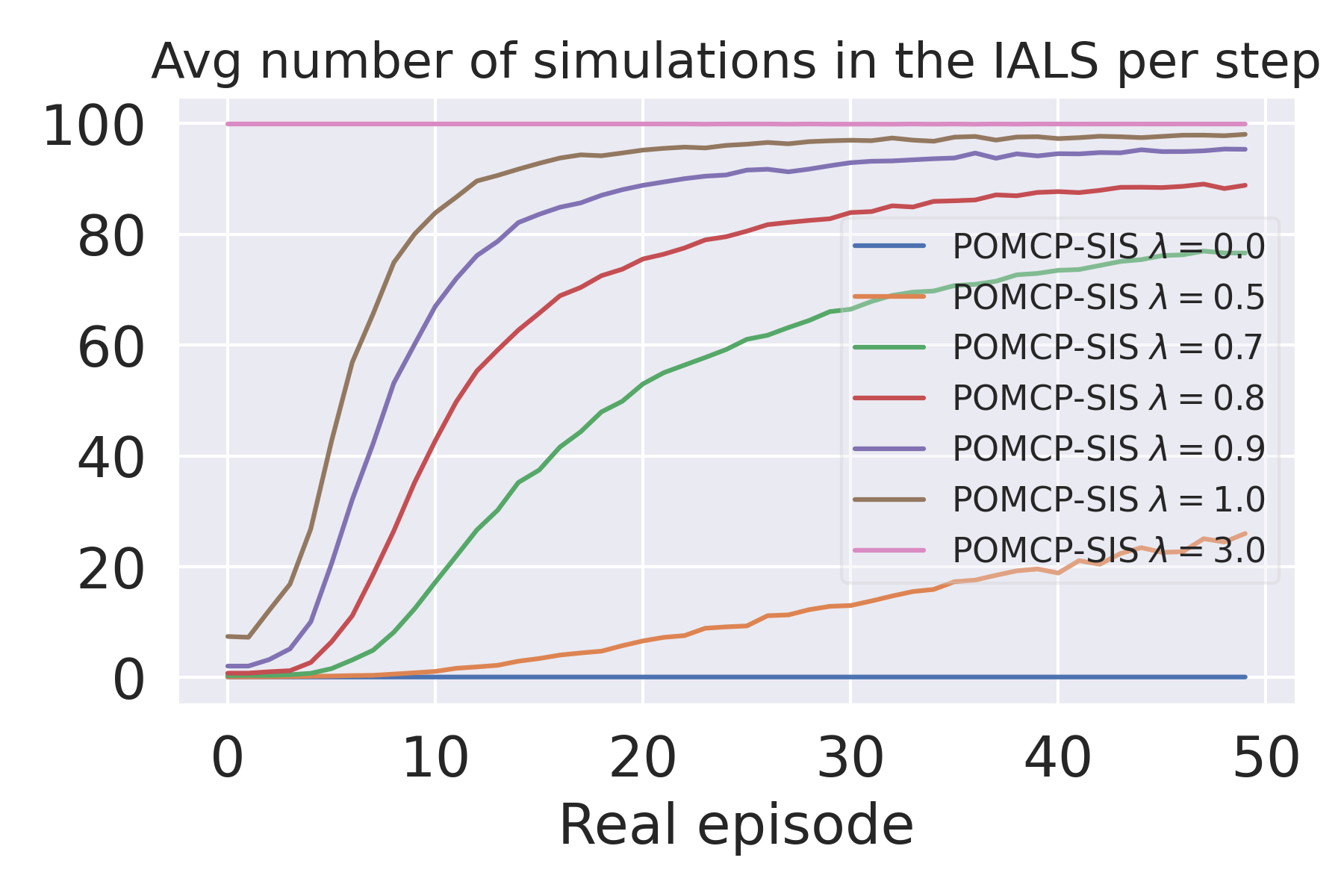}
        \includegraphics[width=0.32\textwidth]{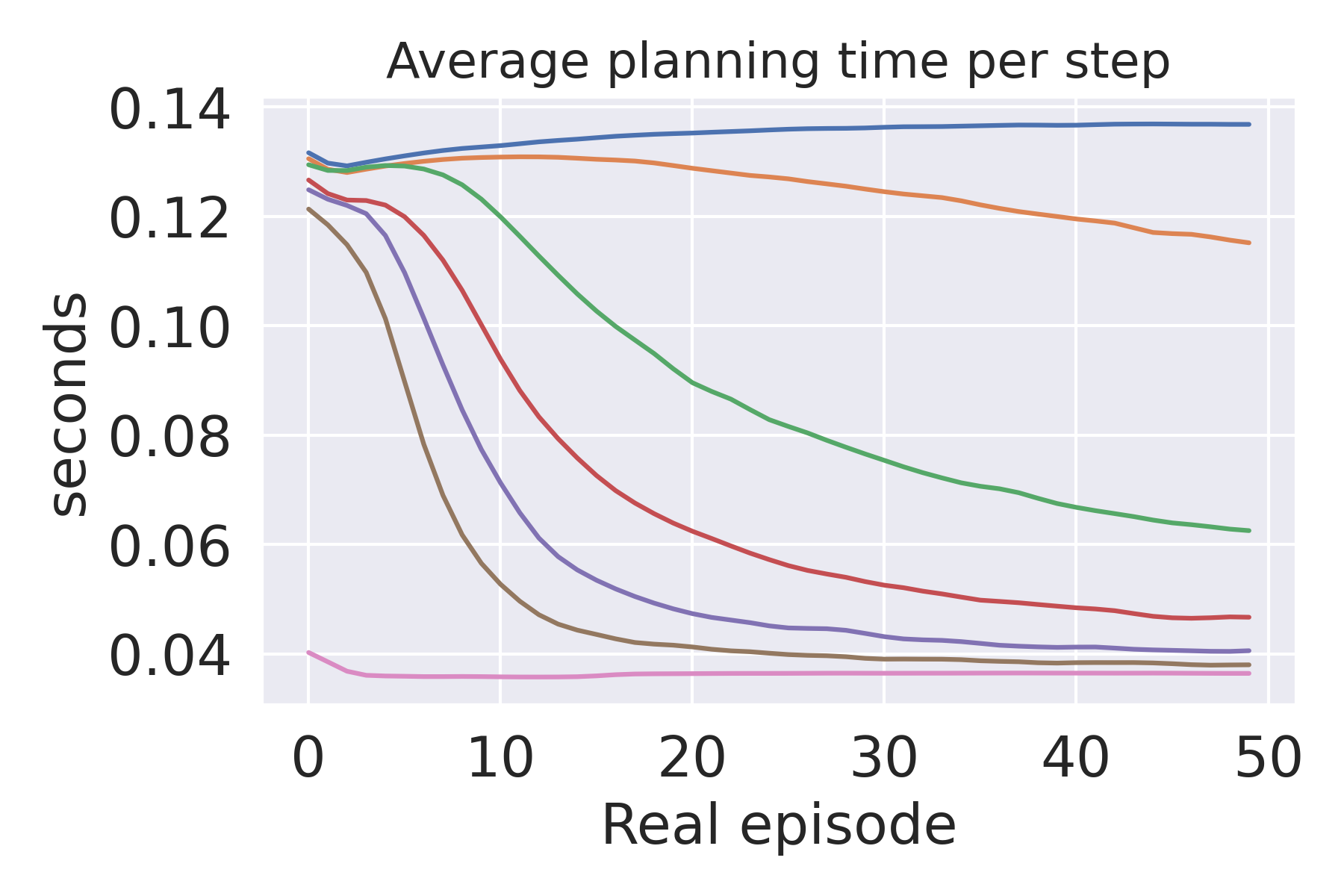}
        \includegraphics[width=0.32\textwidth]{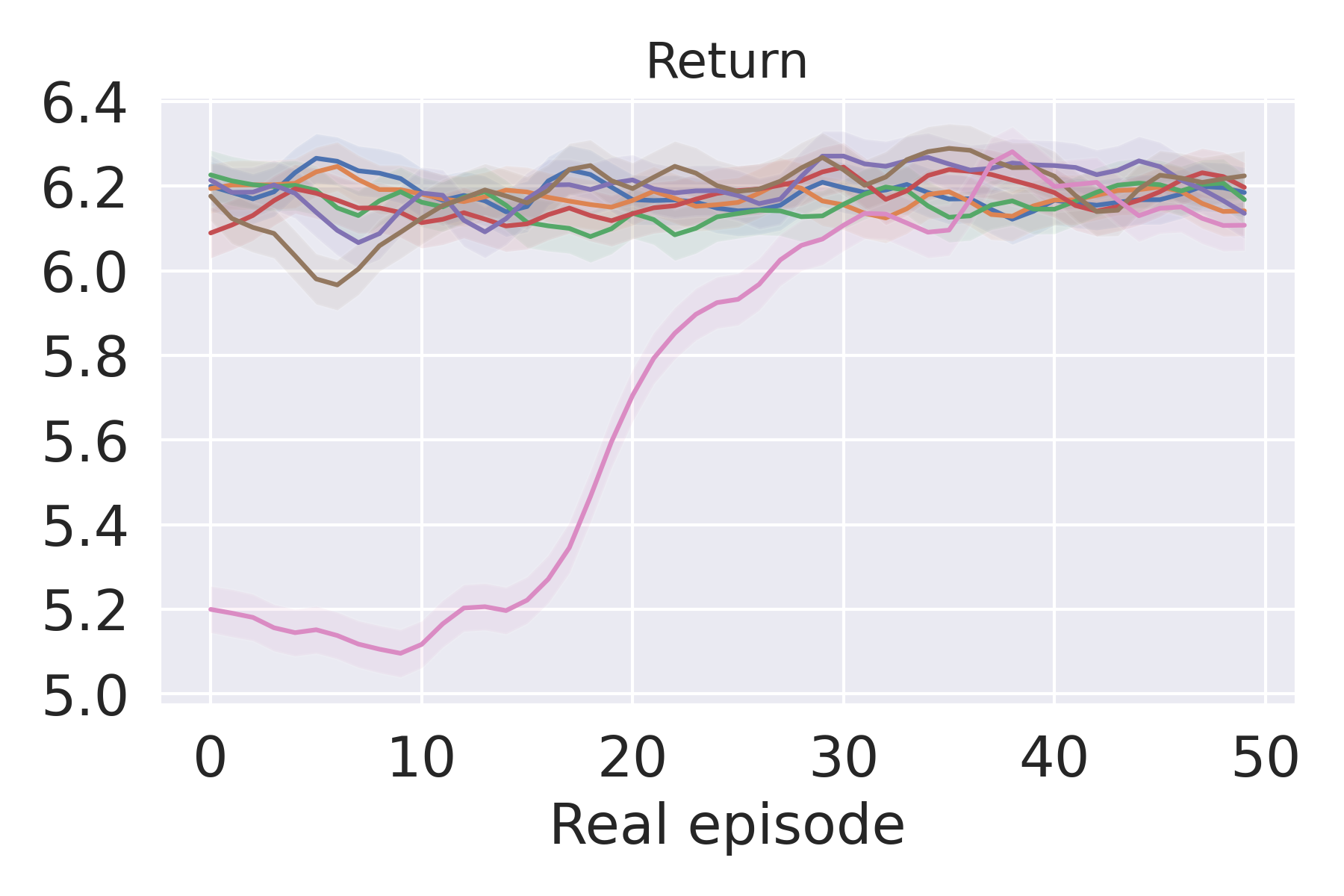}
        \caption{$c^{\mathtt{meta}}=0.3$}
    \end{subfigure}
    \begin{subfigure}[b]{\textwidth}
        \includegraphics[width=0.32\textwidth]{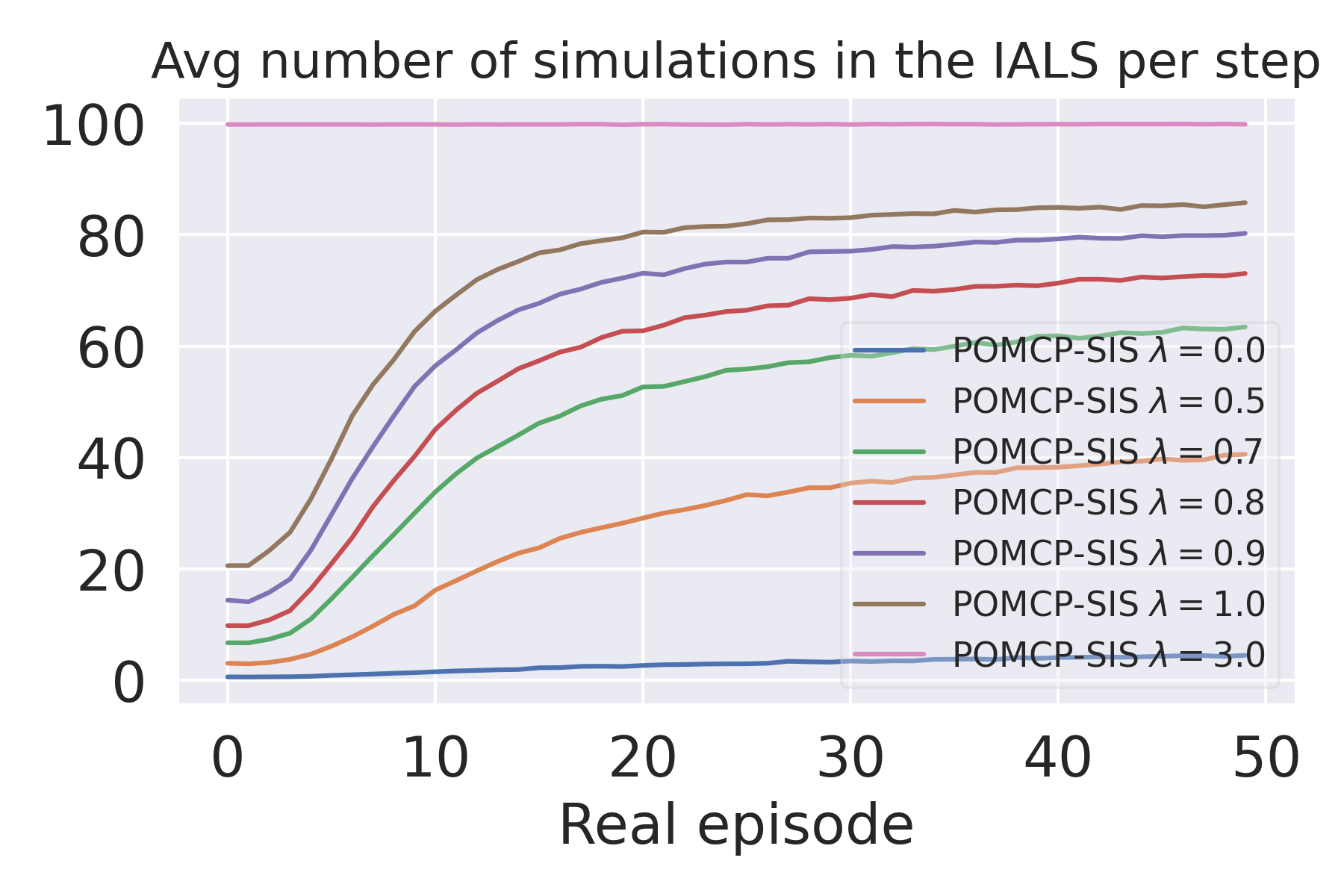}
        \includegraphics[width=0.32\textwidth]{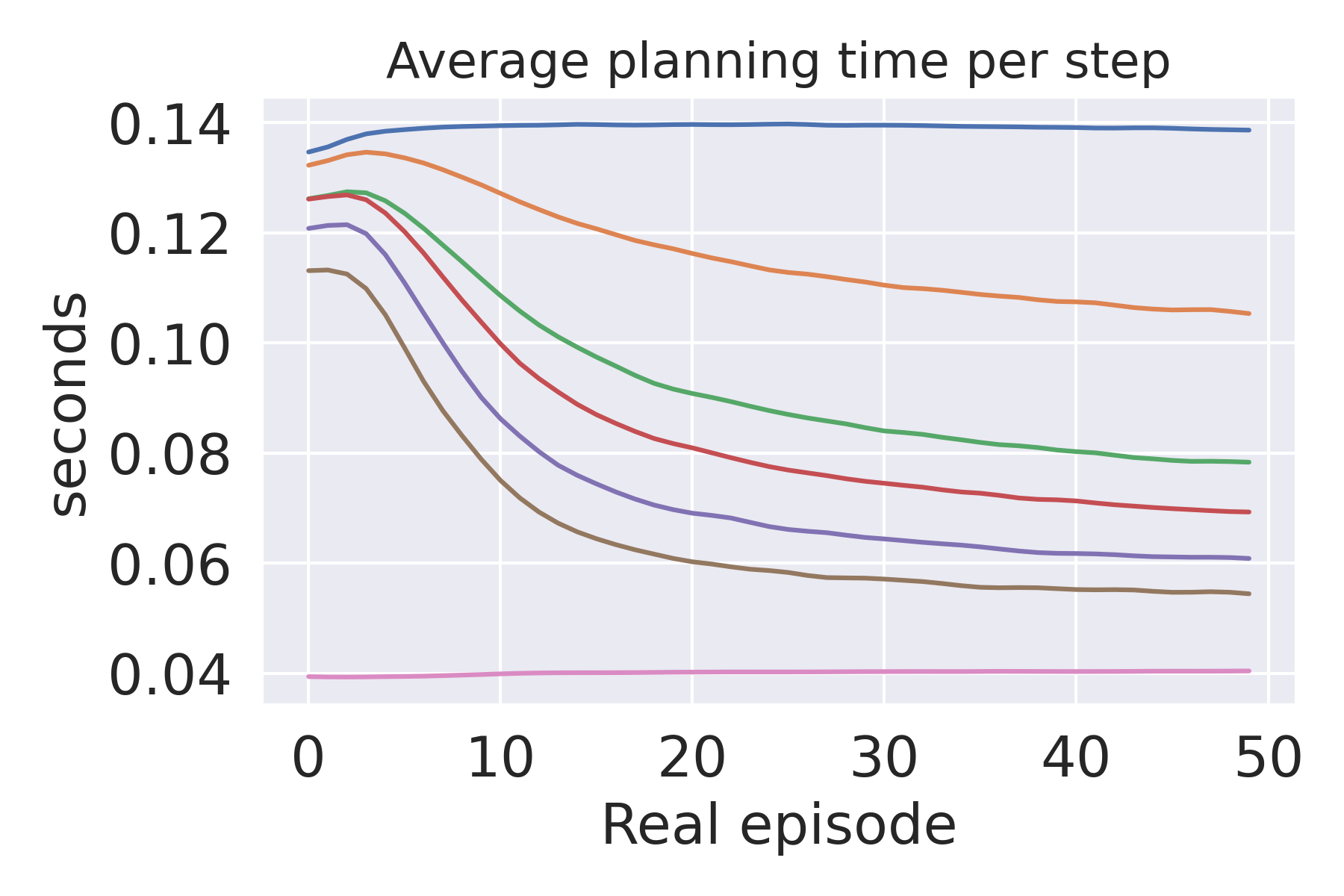}
        \includegraphics[width=0.32\textwidth]{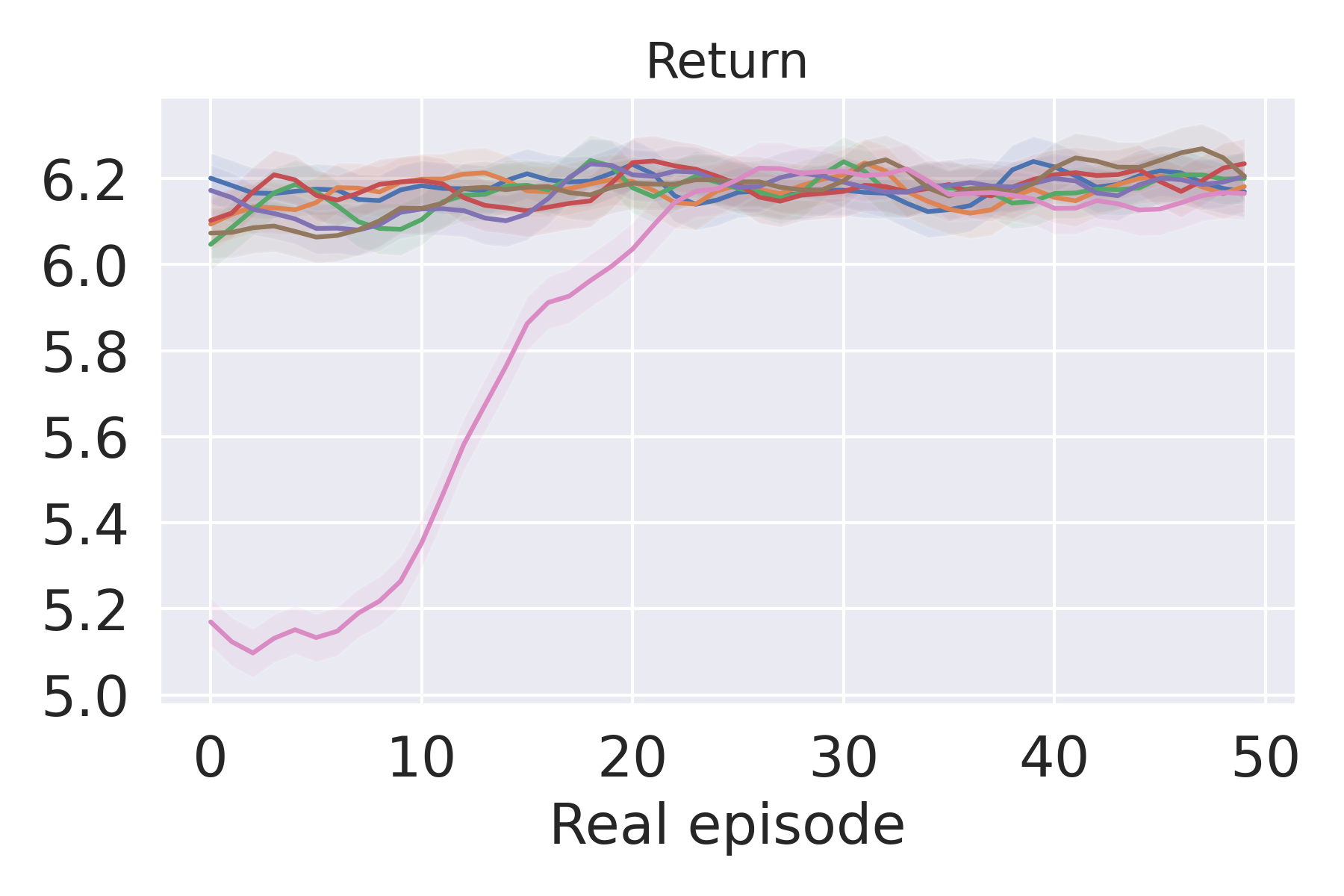}
        \caption{$c^{\mathtt{meta}}=1.0$}
    \end{subfigure}
    \begin{subfigure}[b]{\textwidth}
        \includegraphics[width=0.32\textwidth]{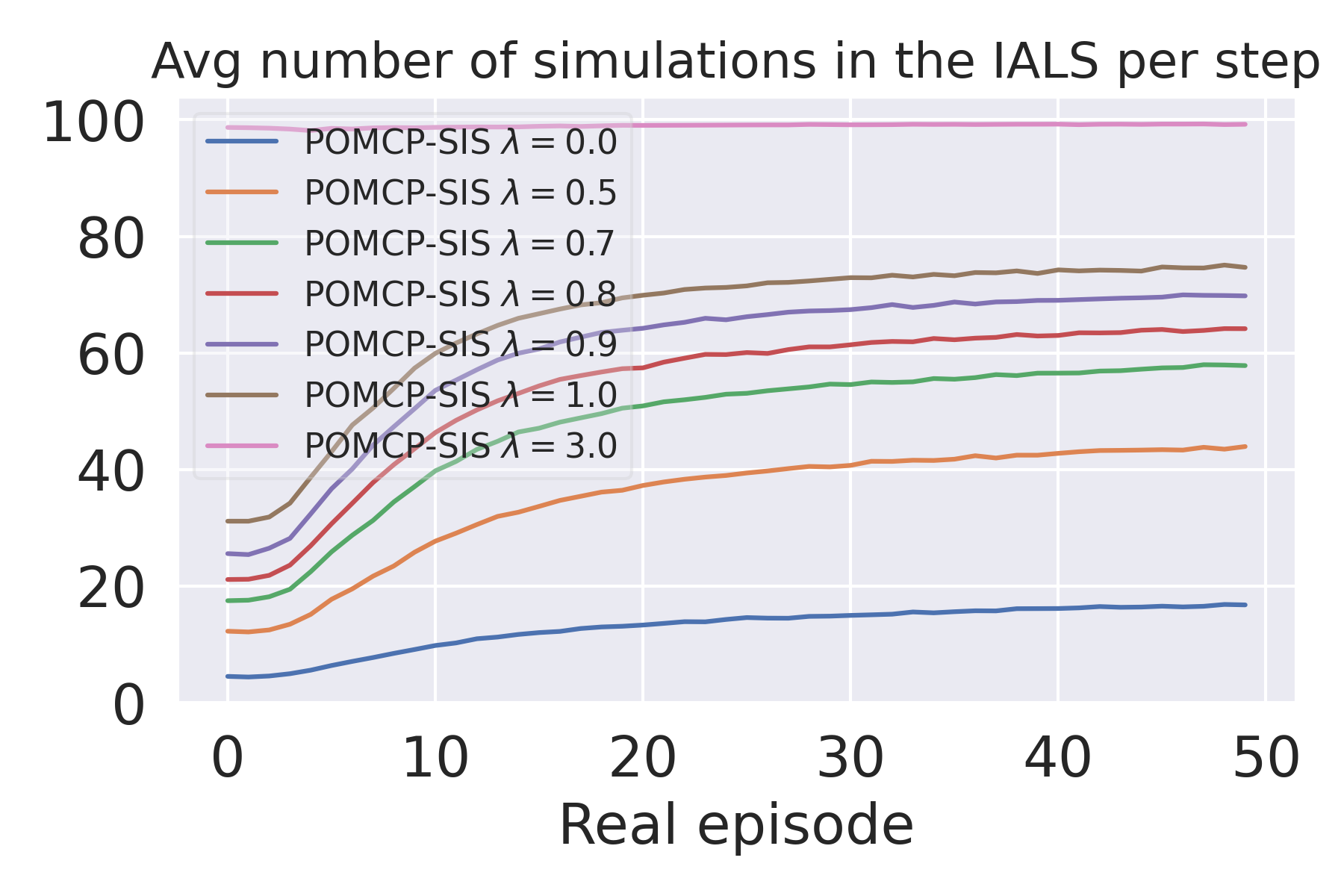}
        \includegraphics[width=0.32\textwidth]{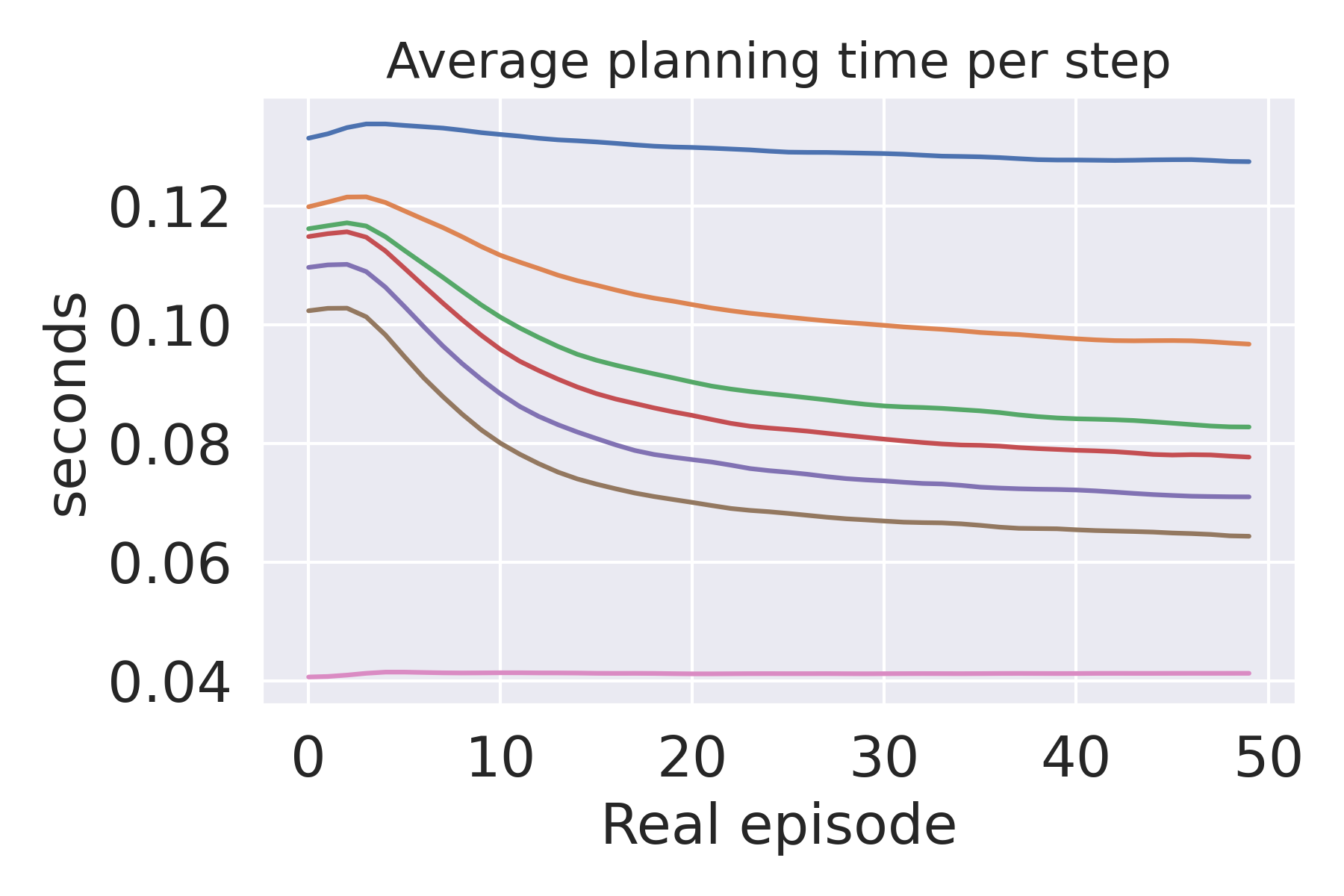}
        \includegraphics[width=0.32\textwidth]{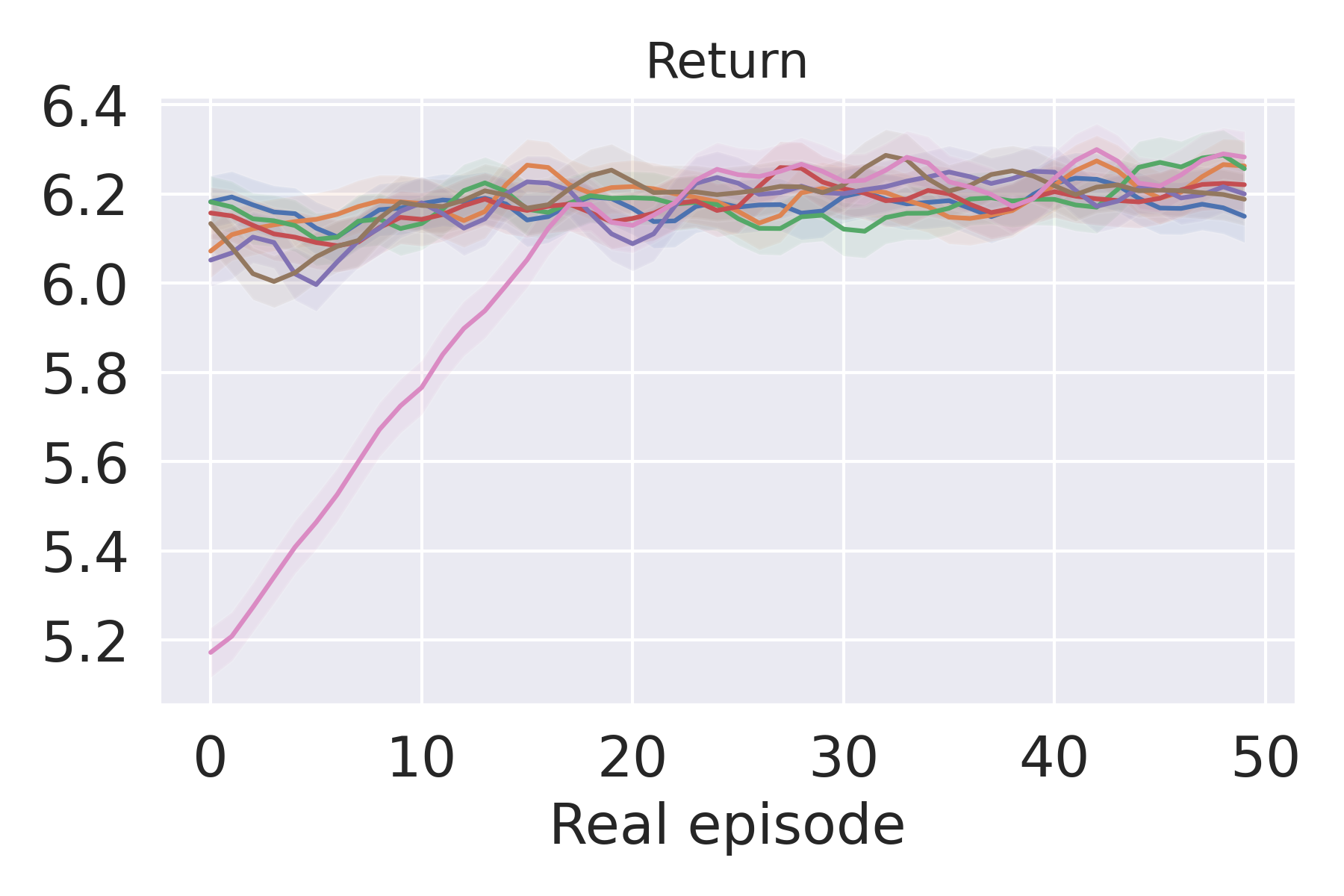}
        \caption{$c^{\mathtt{meta}}=2.0$}
    \end{subfigure}
    \caption{Simulation controlled planning experiments in the grab a chair domain with different values of the meta exploration constant.}
        \label{fig:ablation-mec}
\end{figure*}

}


\end{document}